# Evolving the pulmonary nodules diagnosis from classical approaches to deep learning-aided decision support: three decades' development course and future prospect

Bo Liu [a,1,*], Wenhao Chi [a,f,1], Xinran Li [e], Peng Li [a],

Wenhua Liang [b,d,1], Haiping Liu [c,d,1], Wei Wang [b,d,1], Jianxing He [b,d,*]

[a] *Academy of Mathematics and Systems Science, Chinese Academy of Sciences, Beijing, China*

[b] *Department of Thoracic Surgery and Oncology, The First Affiliated Hospital of Guangzhou Medical University, Guangzhou, China*

[c] *PET/CT Center, The First Affiliated Hospital of Guangzhou Medical University, Guangzhou, China*

[d] *China State Key Laboratory of Respiratory Disease, Guangzhou, China*

[e] *Department of Mathematics, University of Wisconsin-Madison, Madison, WI 53706 USA*

[f] *University of Chinese Academy of Sciences, Beijing, China*

**Abstract**

**Purpose** Lung cancer is the commonest cause of cancer deaths worldwide, and its mortality can be reduced significantly by performing early diagnosis and screening. Since the 1960s, driven by the pressing needs to accurately and effectively interpret the massive volume of chest images generated daily, computer-assisted diagnosis of pulmonary nodule has opened up new opportunities to relax the limitation from physicians' subjectivity, experiences and fatigue. And the fair access to the reliable and affordable computer-assisted diagnosis will fight the inequalities in incidence and mortality between populations. It has been witnessed that significant and remarkable advances have been achieved since the 1980s, and consistent endeavors have been exerted to deal with the grand challenges on how to accurately detect the pulmonary nodules with high sensitivity at low false-positives rate as well as on how to precisely differentiate between benign and malignant nodules. There is a lack of comprehensive examination of the techniques' development which is evolving the pulmonary nodules diagnosis from classical approaches to machine learning-assisted decision support. The main goal of this investigation is to provide a comprehensive state-of-the-art review of the computer-assisted nodules detection and benign–malignant classification techniques developed over 3 decades, which have evolved from the complicated ad hoc analysis pipeline of conventional approaches to the simplified seamlessly integrated deep learning techniques. This review also identifies challenges and highlights opportunities for future work in learning models, learning algorithms and enhancement schemes for bridging current state to future prospect and satisfying future demand.

[*] Corresponding authors. *E-mail address*: bliu@amss.ac.cn (B. Liu), drjianxing.he@gmail.com (J.X. He)

[1] These authors contributed equally to this work.

**Conclusion** It is the first literature review of the past 30 years' development in computer-assisted diagnosis of lung nodules. The challenges indentified and the research opportunities highlighted in this survey are significant for bridging current state to future prospect and satisfying future demand. The values of multifaceted driving forces and multidisciplinary researches are acknowledged that will make the computer-assisted diagnosis of pulmonary nodules enter into the main stream of clinical medicine, and raise the state-of-the-art clinical applications as well as increase both welfares of physicians and patients. We firmly hold the vision that fair access to the reliable, faithful, and affordable computer-assisted diagnosis for early cancer diagnosis would fight the inequalities in incidence and mortality between populations, and save more lives.

**Abbreviations**

| | |
|---|---|
| *2D* | Two-dimensional |
| *3D* | Three-dimensional |
| *AAPM* | American Association of Physicists in Medicine |
| *AI* | Artificial intelligence |
| *ANN* | Artificial neural network |
| *AUC* | Area under the receiver operating characteristic curve |
| *BPNN* | Back propagation neural network |
| *CAD* | Computer-aided/assisted diagnosis |
| *CANN* | Convolutional autoencoder neural network |
| *CNN* | Convolutional neural network |
| *CT* | Computed tomography |
| *CV* | Curvedness |
| *DBN* | Deep belief network |
| *DLCST* | Danish Lung Cancer Screening Trial |
| *DSB* | Data science bowl |
| *ELCAP* | Early Lung Cancer Action Project |
| *FOM* | Figure of merit |
| *EPANN* | Evolved Plastic Artificial Neural Network |
| *EU* | European Union |
| *GDPR* | General data protection and regulation |
| *GLCM* | Gray level co-occurrence matrix |
| *GPU* | Graphics processing unit |
| *HIST* | Histogram analysis |
| *IARC* | International Agency for Research on Cancer |
| *IDRI* | Image Database Resource Initiative |
| *JAFROC* | Jackknife alternative free-response |
| *KNN* | *k*-Nearest neighbors |
| *LASSO* | Least absolute shrinkage and selection operator |
| *LBP* | Local binary pattern |
| *LDCT* | Low-dose computed tomography |
| *LIDC* | Lung Image Database Consortium |
| *LUNA16* | LUng Nodule Analysis 2016 |

*MC-CNN* Multi-crop CNN

*MILD* Multicentric Italian Lung Detection

*MTANN* Massive training artificial neural network

*MTL* Multi-task learning

*NCDs* Noncommunicable Diseases

*NCI* National Cancer Institute

*NLST* National Lung Cancer Screening Trial

*NN* Neural network

*PCF* Prevent Cancer Foundation

*PET* Positron emission tomography

*RBM* Restricted Boltzmann machines

*ROC* Receiver operating characteristic

*SDAE* Stacked denoising autoencoder

*SIFT* Scale invariant feature transform

*SI* Shape index

*SPIE* International Society for Optics and Photonics

*SUV* Standardized uptake value

*SVM* Support vector machine

*TPE* Tree Parzen estimator

*WHO* World Health Organization

**Acknowledgements**

We thank the anonymous referees and editors for their constructive comments in advance on earlier drafts of this manuscript. We thank Emeritus Prof. M.A. Keyzer (SOW-VU, Vrije Universiteit Amsterdam, The Netherlands), Yihui Jin, Ling Wang, Xiaokui Yang (Tsinghua University), Jikun Huang (Peking University), Yubao Guan, Ying Chen, Danxia Huang (China State Key Laboratory of Respiratory Disease and The First Affiliated Hospital of Guangzhou Medical University), Jun Huang (The Sixth Affiliated Hospital of Sun Yat-Sen University), Xiao Han (The Fourth Medical College of Peking University), Rong Zhu (Department of Statistics, Columbia University), Ce Xu, Yi Zhang, Guoxue Wei (National Development and Reform Commission of China), Zhiguang Ren (National Natural Science Foundation of China), Shouyang Wang, Xiaoyi Feng, Yixin Yang, Zhengyang Li, Xin Lyu, Mengchen Ji, Xin Yun, Wenzhe Duan, Keyao Wang (Academy of Mathematics and Systems Science, Chinese Academy of Sciences), Ying Liu (Beihang University) for their insightful discussions and comments on earlier versions of this paper.

## I. Introduction

Lung cancer continues to be the leading cause of cancer mortality worldwide, approximately accounting for nearly one in five cancer deaths, and lung cancer deaths are on the rise (The Lancet 2018). In 2012, lung cancer was responsible for 1.59 million deaths, accounting for 19.4% of the total death from cancer (8.2 million) (Ferlay et al. 2013). In 2015, there were 1.69 million deaths from lung cancer, accounting for 19.2% of the total death from cancer (8.8 million) (World Health Organization 2017). According to the latest released GLOBOCAN 2018 estimates by the International Agency for Research on Cancer (IARC), in 2018 there will be an estimated 1.76 million lung cancer deaths (18.4% of the total cancer deaths) followed by colorectal cancer (9.2%) (Bray et al. 2018). Figure 1 illustrates the estimated number of deaths of the top 10 cancers worldwide in 2018 (Bray et al. 2018; Ferlay et al. 2018).

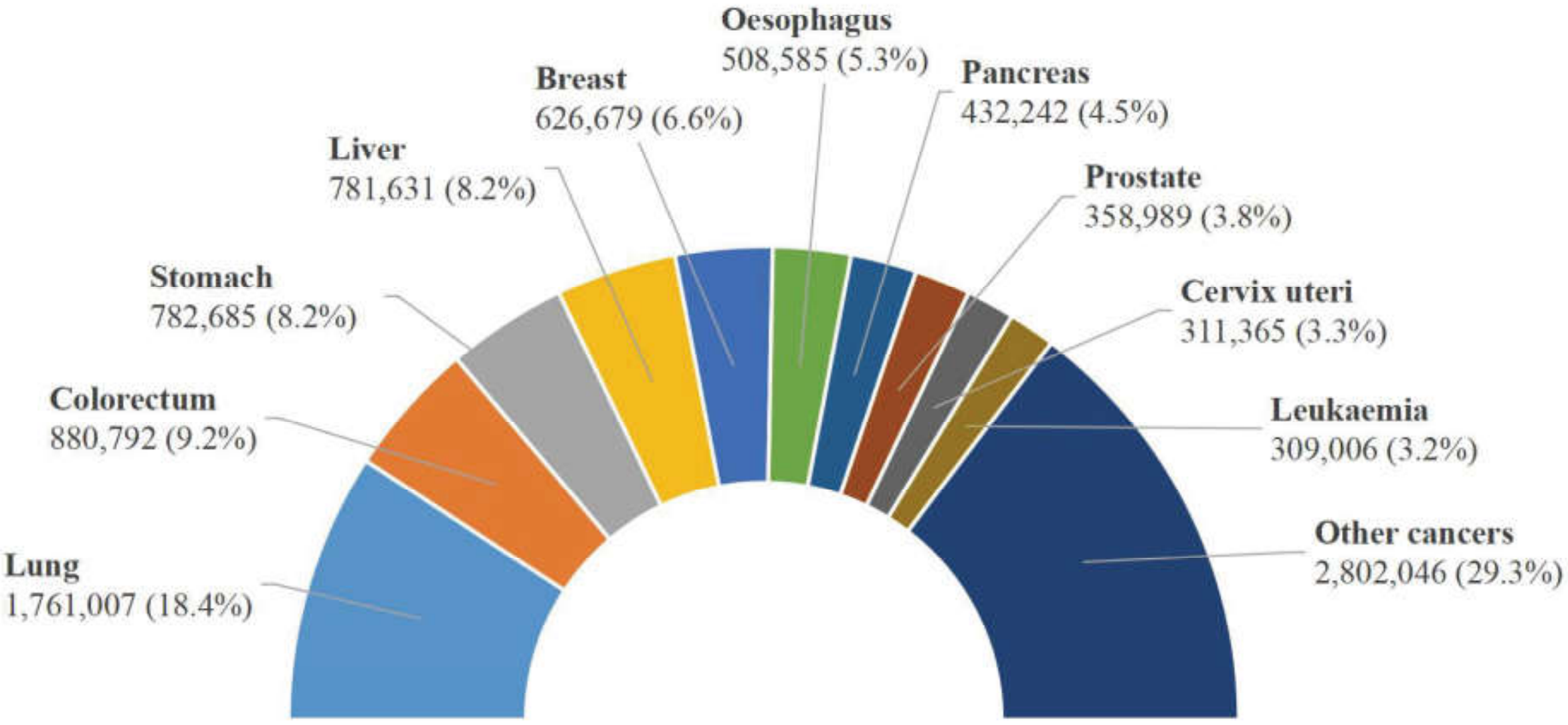

**Fig. 1** Estimated number of deaths of the top 10 cancers worldwide in 2018 from GLOBOCAN (Ferlay et al. 2018)

It is estimated that nearly one-half of lung cancer deaths in the world will occur in China (39.5%) and the United States (8.7%) in 2018 (Bray et al. 2018), shown in Fig. 2. In the United States, 1.7 million new cancer cases and 0.61 million cancer deaths are projected to occur in 2018, among which the lung cancer takes the lead, with approximately 0.23 million new cases (15% of the total) and 0.15 million deaths (25% of the total) (American Cancer Society 2018; Siegel et al. 2018). China now also has very urgent agenda to address the enormous challenges arising from increasing cancer incidence and mortality and China ranks in the top in the incidence and mortality of lung cancer among all countries/regions (Chen et al. 2015; Chen et al. 2016; She et al. 2013). Since 2008, lung cancer took the place of the liver cancer as the leading cause of cancer death in China (She et al. 2013). In 2015, an estimated 4.29 million new cancer cases and 2.81 million cancer deaths occurred in China, with 0.73 million new cases (17% of the total) and 0.61 million deaths (22% of the total) due to lung cancer (Chen et al. 2016). In 2018, there are an estimated 0.77 million new cases and 0.69 million deaths from lung cancer (Bray et al. 2018), which are higher than the numbers of 2015.

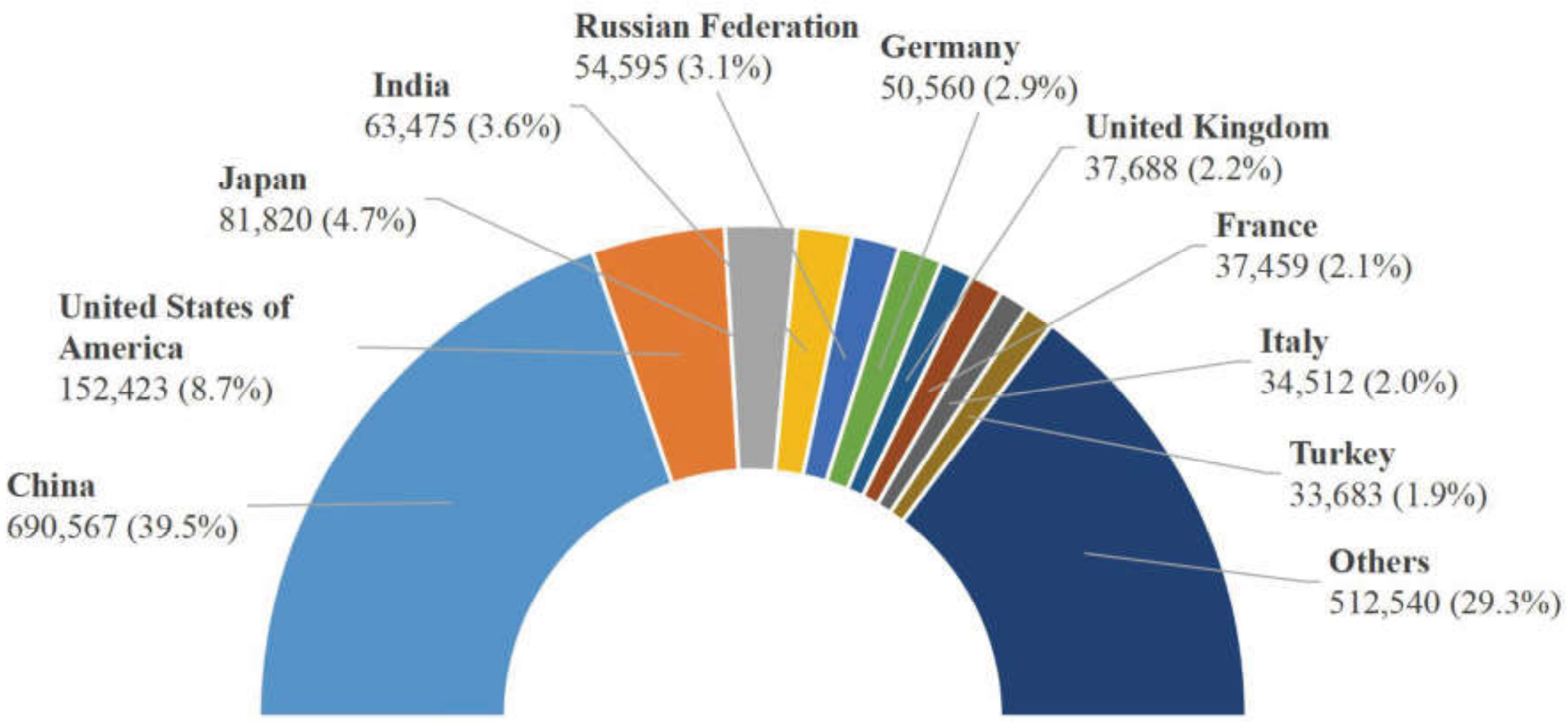

**Fig. 2** Estimated number of deaths of the lung cancer of the 10 highest countries/regions in 2018 from GLOBOCAN (Ferlay et al. 2018)

Lung cancer is continuing to affect considerable portion of the world's population, especially to the vulnerable population in the low- and middle- income countries. Approximately 70% of cancer deaths occur in the low- and middle- income countries (World Health Organization 2017) and impose an enormous economic burden on the patients and fragile social safety net (American Cancer Society 2018; National Health and Family Planning Commission of PRC 2017; Stewart and Wild 2014). In 2013, the Global Action Plan for the Prevention and Control of Noncommunicable Diseases (NCDs) 2013-2020 was initiated by World Health Organization (WHO), aiming at a 25% reduction in premature mortality from four types of NCDs (i.e., cancer, cardiovascular diseases, diabetes and chronic respiratory diseases) by 2025 (World Health

Organization 2013). To achieve the goal set by the action plan, the WHO and IARC are collaborating with partners to take intervening actions on improved cancer prevention, early diagnosis, screening and treatment.

Lung cancer mortality can be reduced if cases are detected through early diagnosis and screening and treated early (Swensen et al. 2003). And it is well established that the computer-assisted/aided diagnosis containing transferred expertise's knowledge could support faithful early diagnostic decision to complement/enhance the competence of physicians. In 1960s, Lodwick et al. initiated computer-assisted diagnosis (CAD) which used the computer as the aid in radiologic diagnosis (Lodwick 1966), and conducted CAD for lung cancer (Lodwick et al. 1963). The images on roentgenograms were converted into quantitative form which could be recognized and processed by the computer. This pioneering work made possible the exploration of the use of the computer as an aid in radiologic diagnosis, as well as evaluation of the operability and predicting survival rate. Even though this investigation was not focused on detection or classification of pulmonary nodules, they opened up the direction of computer-assisted diagnosis for medical purpose via converting medical images into quantitative features, which paved the way for the later development of early diagnosis of lung cancer. Since then, computer-assisted diagnosis of cancer has received considerable theoretical, computational, and empirical research work. To date, a large amount of computer-assisted diagnosis methods have been developed for detection and classification of lung/pulmonary nodules in different formats of images, e.g., chest radiographs, computed tomography (CT), and positron emission tomography (PET) (Giger et al. 1990; Giger et al. 1988; Greenspan et al. 2016; Gurney 1993; Lampeter and Wandtke 1986; Miotto et al. 2017; Stitik et al. 1985; van Ginneken 2017), which made attempts to provide efficient and effective technical assistance to physicians.

During the last 5 decades, researchers and practitioners in multidisciplinary fields not limited in radiology, radiomics, clinical medicine, computer sciences, mathematics and optimization have been witnessing the pressing needs for new efforts in dealing with the grand challenges arising from computer-assisted decision support for pulmonary nodules diagnosis (Morris et al. 2018). This is motivated, first by the fact that lung cancer mortality could be reduced significantly by performing early diagnosis and annual low-dose lung CT screening (Meyer et al. 2017), for example, the US National Lung Screening Trial (NLST) (The National Lung Screening Trial Research Team 2011), EU position statement (Oudkerk et al. 2017), Dutch-Belgian Randomized Lung Cancer Screening Trial (NELSON) (Heuvelmans et al. 2017; Horeweg et al. 2014), Multicentric Italian Lung Detection (MILD) (Pastorino et al. 2012; Silva et al. 2018; Sverzellati et al. 2016), and Danish Lung Cancer Screening Trial (DLCST) (Pedersen et al. 2009). To improve the survival rate, the core challenges in lung cancer early diagnosis and screening are imposed on the accurate detection (high sensitivity with low false-positives rate) and precise classification between benign and malignant nodules especially at early stage; second by the fact that pulmonary images are continuously generated in an unprecedented amount daily, leading to grand challenges of high volume data, to avoid the physicians' subjectivity, less experiences and fatigue when interpreting the medical images, efficient and effective automatic diagnosis and assessment of pulmonary nodules is indispensable; third by the fact that computer-assisted decision support system, which is kind of knowledge-based end-to-end technologies, could be rapidly disseminated and deployed to the vulnerable population in less-developed countries or regions with effectively transferred expertise's knowledge to support faithful diagnostic decision; last but not least by the fact that the emerging deep learning techniques (e.g., deep convolutional neural networks, semi/un-supervised learning, transfer leaning, continual learning, and shared learning) would accelerate the development of computer-assisted diagnosis to make better and safer decision in precedent/unprecedented cases to enhance the competence of physicians, even in some cases on the behalf of the physicians (Stead 2018). The value of multifaceted driving forces will make the computer-assisted diagnosis enter into the main stream of clinical medicine and increase both welfares of physicians and patients.

To date, several literature reviews on the advances and progresses of automatic diagnosis of pulmonary nodules have been reported (Greenspan et al. 2016; Valente et al. 2016; van Ginneken 2017). In 2016 Greenspan et al. (2016) organized a special issue of IEEE Transactions on Medical Imaging on deep learning in medical imaging, and presented recent achievements of deep-learning applications to a variety of medical tasks, including lesion detection, segmentation, shape modeling, and registration. Among the 18 selected papers in this special issue, pulmonary nodule detection in CT images was well addressed by Setio et al. (2016). van Ginneken (2017) illustrated a few research work focusing on how to apply rule-based, machine learning, convolutional neural network-based deep learning-techniques to important issues in analysis of chest imaging, from rib detection and suppression in chest radiographs, fissure extraction, airway segmentation, nodule detection, to nodule classification and characterization in CT. Valente et al. (2016) put forward the first review devoted exclusively to automated 3D techniques for the detection of pulmonary nodules in CT images. Their research covered the published works up to December 2014. Yang et al. exclusively surveyed the deep learning techniques for pulmonary nodules diagnosis during 2015-2017 (Yang et al. 2018). Razzak et al. (2018) discussed the state-of-the-art deep learning architecture and its optimization for medical image segmentation and classification in which pulmonary nodules were not discussed.

Furthermore, to provide fair comparisons between various nodule detection and classification algorithms, LUNGx Challenge (Armato et al. 2016; Armato et al. 2015), Data Science Bowl (DSB) 2017, and LUng Nodule Analysis 2016 (LUNA16) challenge (Setio et al. 2017) were conducted.

Significant and remarkable advances have been achieved for pulmonary nodules diagnosis after 30 years' development, and consistent endeavors have been exerted to accurately detect the pulmonary nodules with high sensitivity as well as on to precisely differentiate between benign and malignant nodules. There is a lack of comprehensive examination of the techniques' development which is evolving the pulmonary nodules diagnosis (e.g., nodule detection and benign-malignant classification) from classical approaches to machine learning-assisted decision support. We attempt to fill in the gap by exclusively focusing on computer-assisted diagnosis of pulmonary nodules and provide a comprehensive state-of-the-art review of the computer-assisted nodules detection and benign-malignant classification techniques developed over 3 decades, and we show the methodological evolving path from the complicated ad hoc analysis pipeline of conventional approaches to the simplified seamlessly integrated deep learning techniques. The review covers extensively the state-of-the-art analysis framework, analysis methods (pre-processing, matching, feature extraction, fine filter, and classifier), public data sets, and comparative results. We searched on arXiv, Cochrane, Elsevier, Google Scholar, IEEE, MICCAI, PubMed, RSNA, SPIE Medical Imaging, Springer, and Web of Science with keywords "lung/pulmonary nodule", "lung cancer", "computer-assisted/aided", "detection", "classification", "diagnosis", "benign", "malignant" and their combinations. Within these searching results which were inclusive enough, a grand picture of the evolving path of the area was uncovered by eliminating huge body of un(less-)related literature. During this tedious screening process, some related research work may be left out unintentionally. As far as the authors know, it is the first literature review of the past 30 years' development in computer-assisted diagnosis of lung nodules. We hope that we provide a holistic and comprehensive research skeleton of success stories of pulmonary nodules diagnosis.

We are now witnessing the emerging of artificial intelligence (AI), especially deep learning as promising decision supporting approaches to automatically analyze medical images for different clinical diagnosing purposes (Bibault et al. 2016; Esteva et al. 2017; Greenspan et al. 2016; Gulshan et al. 2016; Hinton 2018; LeCun et al. 2015; Litjens et al. 2017; Miotto et al. 2017; Naylor 2018; Ravì et al. 2017; Schmidhuber 2015; Stead 2018; Zhou et al. 2017). Even recently, the U.S. Food and Drug Administration permitted the marketing of the first artificial intelligence-based medical device to detect certain diabetic retinopathy (Food and Drug Administration 2018). However, the promising AI/deep learning techniques raise daunting challenges, and tremendous technical barriers need to be crossed. The second aim of this investigation is to identify several open questions on data scarcity, training efficacy, and diagnostic accuracy. And we highlight future research opportunities in (1) learning models (e.g., semi/un-supervised learning, transfer learning, multi-task learning) and data contribution mechanism (e.g., crowd sourcing) for making up for the lack of data, (2) learning algorithms (e.g., optimal sampling strategy and optimization techniques) for speeding up training process which is often time-consuming and slow-convergence, and (3) enhancement schemes (explainable and robust decision, continual learning and shared learning, and integration with big data) for improving diagnostic accuracy. The review summarizes the achievements accumulated in literature and practice, and highlights the directions to bridge current state to future prospect. The value of tackling of the challenges will offer a way for putting these ideas into practice to raise the state-of-the-art clinical applications for better diagnosis as well as to increase both welfares of physicians and patients.

## II. Computer-assisted pulmonary nodule detection

Detection of the pulmonary nodule traditionally involves two consecutive procedures, i.e., matching and fine filtering. The illustrative framework of computer-assisted detection of pulmonary nodules is depicted in Fig. 3. The purpose of matching is to detect the lung nodules or the suspected regions of interest with high sensitivity, while the fine filtering is to eliminate the negative nodules to achieve low false positive rates. However, the high sensitivity and low false-positives rate are two paradoxical concerns. High sensitivity usually leads to high number of false positives. Therefore, to balance the two paradoxical concerns is the core challenge for accurate detection.

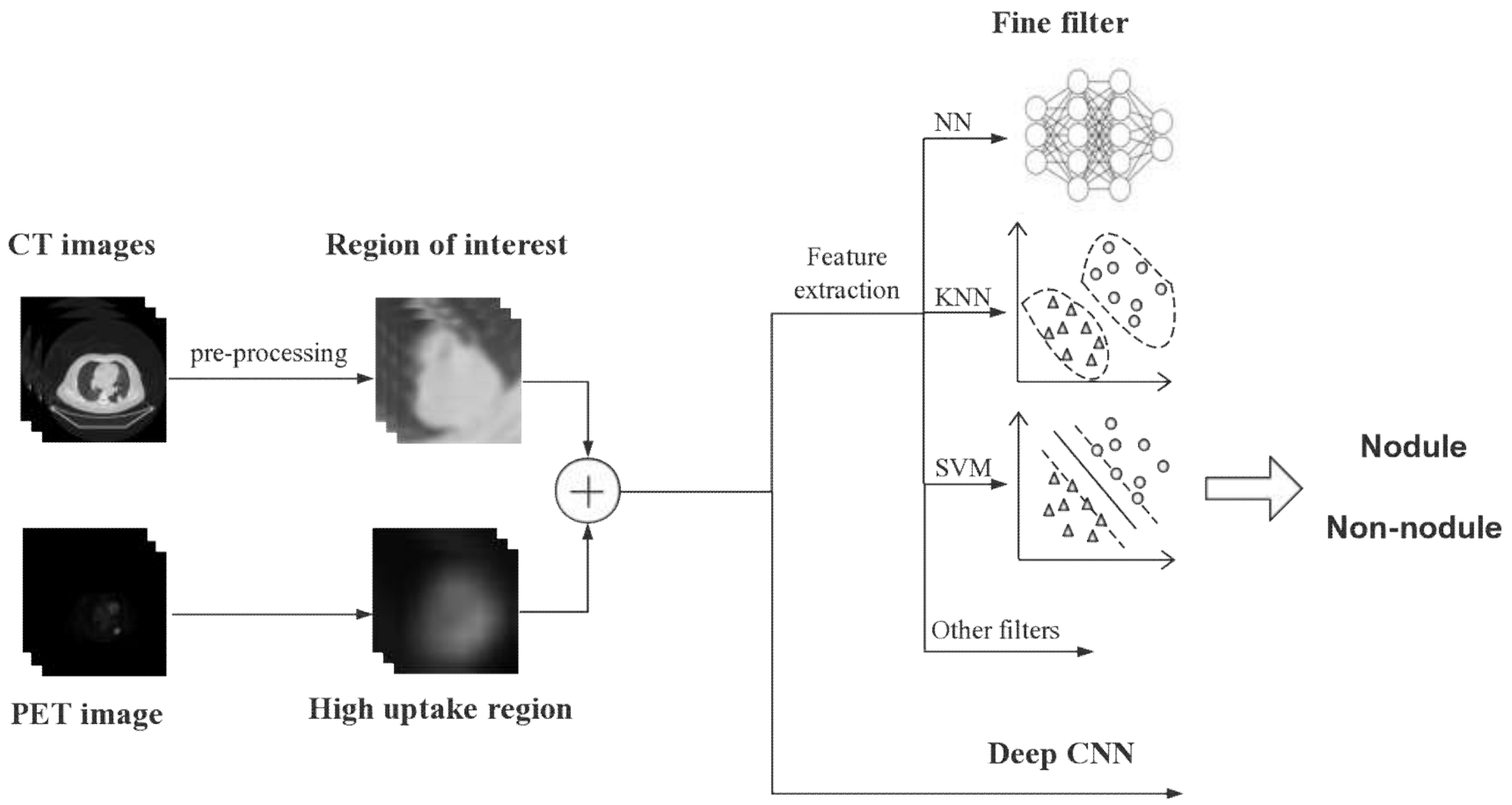

**Fig. 3** Illustrative framework of computer-assisted detection of pulmonary nodules

Based on nodule matching and fine filtering paradigms, we roughly divide the methodologies into two basic categories—classical detection methods and deep learning aided methods. The classical detection methods require the well defined explicit features of a disease pattern, determine the suspicious regions of interest by maximizing the matching rate between the featured profiles and the suspected areas, and further reduce the false positives by elimination of false nodules from the nodule candidates via performing fine filtering equipped with well defined features of true nodules. Unlike the ad hoc analysis pipeline of classical detection methods, the deep learning aided methods can internally extract the implicit features for disease patterns and are capable of discerning among different patterns, which suggests that exploitation of features and tuning of performance are seamlessly integrated in one framework of deep learning.

In this section, we analyze the studies dedicated to pulmonary nodule detection with the classical nodule detection techniques and the deep learning techniques. Table 1 summarizes the detection techniques.

**Table 1** Detection techniques for pulmonary nodules

| *Authors* | *Year* | *Methods* | *Effects* | *Data set Size* | *Data Dimension* | *Image Format /Data Source* |
|---|---|---|---|---|---|---|
| Giger et al. (1988) | 1988 | (1) Preprocessing: difference image approach<br>(2) Feature extraction: thresholding and circularity calculation | Sensitivity of 80% with 2.7 false positives per image | 6 images | 2D | Chest radiographs |
| Giger et al. (1990) | 1990 | (1) Feature extraction: morphological operation<br>(2) Nonlinear filters: erosion and dilation | Sensitivity of 57.6% with 3.7 false positives per image | 60 images | 2D | Chest radiographs |
| Lo et al. (1993a) | 1993 | (1) Preprocessing: thresholding evaluation and background subtraction<br>(2) Matching: sphere profile matching algorithm<br>(3) Fine filter: backpropagation neural network | AUC of 0.782 | 30 images | 2D | Chest radiographs |
| Lo et al. (1993b) | 1993 | (1) Matching: boundary detection and sphere profile matching algorithms<br>(2) Fine filter: convolutional neural network | AUC of 0.88;<br>Sensitivity of 80% with 2.6 false positives per image | 5 images for training + 21 images for testing | 2D | Chest radiographs |
| Lo et al. (1998) | 1998 | Associative learning neural networks | Sensitivity of 80% with 2.5 false positives per image | 62 images | 2D | Chest radiographs |
| Suzuki et al. (2003) | 2003 | Fine filter: multiple massive training artificial neural network | Sensitivity of 80.3% at an average 4.8 false positives per scan | 101 scans | 2D | Chest CT |
| Murphy et al. (2009) | 2009 | (1) Matching: local image features of shape index and curvedness<br>(2) Fine filter: two successive *k*-nearest neighbors classifiers | Sensitivity of 80.0% / 72.4% / 77.7% with an average 4.2 / 4.0 / 4.2 false positives per scan for database A / B / C, respectively | 1535 scans in database A, 1121 scans in both database B and C (343 scans were also contained in database A) | 2D | Chest CT / NELSON |
| Tan et al. (2011) | 2011 | (1) Matching: nodule segmentation based on nodule and vessel enhancement filters and a divergence feature<br>(2) Fine filter: support vector machines, neural networks, and fixed-topology neural networks | Sensitivity of 87.5% with an average of 4.0 false positives per scan by the neural networks and the fixed-topology neural networks; sensitivity of 83.8% with an average of 4.0 false positives per scan by SVM | 235 scans for training and 125 scans for testing | 2D | Chest CT / LIDC |
| Jacobs et al. (2014) | 2014 | (1) Feature extraction: 128 user defined features including context features, intensity, shape, and texture features<br>(2) Fine filter: linear discriminant classifier plus GentleBoost classifiers | Sensitivity of 80% at an average 1.0 false positive per scan | 209 scans for training and 109 scans for testing | 2D | Chest CT /NELSON |
| Torres et al. (2015) | 2015 | (1) Lung segmentation: ant colony<br>(2) Fine filter: voxel-based feed-forward neural network | Sensitivity of about 80% with 8 false positives per scan | 94 scans for training and 949 scans for testing | 2D | Chest CT / LIDC/IDRI, ITALUNG-CT, and ANODE09 |
| Setio et al. (2015) | 2015 | (1) Feature extraction: 24 features based on intensity, shape, blobness, and spatial context<br>(2) Radial basis support vector | Sensitivities of 94.1% and 98.3% at an average of 1.0 and 4.0 false positives per scan, respectively | 888 scans | 2D | Chest CT / LIDC-IDRI |
| | | machine | | | | |
| Bergtholdt et al. (2016) | 2016 | Fine filter: cascaded support vector machine | Sensitivity of 85.9% with 2.5 false positives per scan | 1018 scans | 2D | Chest CT / LIDC-IDRI |
| Setio et al. (2016) | 2016 | Two-dimensional multi-view deep convolutional neural network | Sensitivities of 85.4% and 90.1% at 1 and 4 false positives per scan | 888 scans | 2D | Chest CT / LIDC-IDRI |
| Teramoto et al. (2016) | 2016 | (1) Preprocessing: active contour filter, predetermined threshold method<br>(2) User defined features: shape and metabolic feature<br>(3) Automatic feature extraction by deep CNN<br>(4) Hybrid classifier: rule-based classifier plus two SVMs | Sensitivity of 90.1% with 4.9 false positives per case | 104 cases | 2D | PET and CT |
| Tajbakhsh and Suzuki (2017) | 2017 | Massive-training artificial neural networks and four deep convolutional neural networks (i.e., AlexNet, rd-CNN, LeNet architecture, and sh-CNN) | AUC of 0.8806 and sensitivity of 100% at 2.7 false positives per patient by MTANNs;<br>AUC of 0.7755 and sensitivity of 100% at 22.7 false positives per patient by CNNs | 38 scans (from 31 patients) | 2D | Chest CT |
| Dou et al. (2017) | 2017 | (1) Three 3-D CNNs to generate the prediction probabilities<br>(2) Weighted linear combination to integrate the three prediction probabilities | All three designs have sensitivity of over 90% with 8.0 false positives per scan | 888 scans | 3D | Volumetric CT / LIDC |
| Setio et al. (2017) | 2017 | Combination of seven nodule detection systems and five false | Sensitivities of 96.9% and 98.2% at 1.0 and 4.0 false positives per scan, | 888 scans | 2D and 3D | Chest CT / LIDC-IDRI |
| | | positive reduction systems | respectively | | | |
| Gupta et al. (2018) | 2018 | (1) Feature extraction: lung segmentation, morphological closing process, rule-based thresholding algorithms<br>(2) Fine Filter: three layered feed-forward neural network | Sensitivity of 85.6% at 8 false positives per scan on LIDC-IDRI (training) data set;<br>sensitivities of 66.3%, 70.4%, and 68.9% at 8 false positives per scan on AAPM–SPIE–LUNGx, ELCAP and PCF (testing) data sets, respectively | 899 scans for training and 153 scans for testing | 2D | Chest CT / LIDC-IDRI, AAPM–SPIE–LUNGx, ELCAP, and PCF |
| Teramoto and Fujita (2018) | 2018 | (1) Matching: cylindrical nodule-enhancement filter<br>(2) Fine filter: rule-based classifier and three support vector machines | Sensitivity of 83% with 5 false positives per case | 50 cases for training with 50 cases for testing | 2D | PET and CT /East Nagoya Imaging Diagnosis Center |
| Jiang et al. (2018) | 2018 | (1) Preprocessing: slope analysis, Frangi filters, and threshold processing<br>(2) Multi-channel deep CNN | Sensitivities of 80.06% and 94% with 4.7 and 15.1 false positives per scan | 1006 scans | 2D | Chest CT / LIDC-IDRI |
| Nam et al. (2019) | 2019 | (1) Three deep CNNs structured by 25 layers and 8 residual connections and trained in a semi- supervised learning manner<br>(2) Results from three Deep CNNS were averaged for the final prediction | For radiograph classification:<br>AUCs of 0.96, 0.92, 0.99, 0.94, and 0.96 on one internal and four external validation data sets;<br>Sensitivities of 79.0%, 91.1%, 71.2%, and 87.6% with specificities of 95%, 98%, 100%, and 93% on four external validation data sets<br>For malignant nodule detection: JAFROC FOMs of 0.852, 0.885, 0.924, 0.831 and 0.880 on one | 42,092 (8625) images for training, 600 (300) images for tuning, and 600 (300) images for validation;<br>181 (119), 182 (123), 181 (111), and 149 (89) images in four external validation data sets [a] | 2D | Chest radiographs / Seoul National University Hospital, Boramae Hospital, National Cancer Center – Korea, and University of California San Francisco Medical Center |
| | | | internal and four external validation data sets;<br>Sensitivities of 69.9%, 82.0%, 69.6%, and 75.0 with 0.34, 0.30, 0.02, and 0.25 false positives per image on four external validation data sets | | | |

[a] Data in parentheses are number of cases with malignant nodules

*A. Classical Pulmonary Nodules Detection Techniques*

*1) Conventional Detection Techniques*: Several conventional computer vision algorithms have been proposed for recognition of the pulmonary nodules, which shed the light on the potential usefulness of computer-aided diagnosis to assist the clinician in locating the lung nodules (Giger et al. 1990; Giger et al. 1988; Lampeter and Wandtke 1986).

Giger et al. (1988) developed thresholding and circularity calculation in detection of nodules in chest radiographs. Their approach involved two steps: first, the difference image approach was applied to eliminate the background anatomic structures from the lung fields; second, feature extraction techniques (e.g., circularity test, size and their variation with threshold level) were investigated to isolate the suspected nodules. The detection scheme was tested on 6 chest radiographs containing 8 peripheral nodules and 2 medial nodules, the true-positive detection rate reached 80% with 16 false-positive detections totally among the six images.

In a successive work by Giger et al. (1990), a morphological processing-based feature extraction scheme was developed to reduce the false-positive detections, which sequentially deployed the nonlinear filters of erosion and dilation to eliminate the camouflaging effect of rib crossings and end-on vessels on nodule recognition. Preliminary test results based on data set consisting of 60 chest radiographs, half of which contained nodules showed that a true-positive detection rate of approximately 57.6% were achieved with an average of 3.7 false-positive detections per chest radiograph. However, compared with the thresholding and circularity calculation in Giger et al. (1988), the computational burden of morphological processing was heavy.

Lo et al. (1993a) proposed a sphere profile matching algorithm to extract and determine the suspected lung nodules. In the algorithm, matching rate was maximized between the potential artificial sphere profiles and the suspected areas. The higher the values of the matching rates obtained in the corresponding extracted image patches, the higher the probability of presence of nodules in this patch. To be noted, the extracted image patches fed to sphere profile matching algorithm were pre-processed for enhancement by the thresholding evaluation and background subtraction proposed in Giger et al. (1988). Some round vessels and rib crossings were misidentified as suspected nodules. Tested on 30 cases including 59 true nodules and 175 non-nodules, the sensitivity of the proposed algorithm was 100%, but the specificity was not high enough.

Though preliminary results of the above conventional image processing scheme showed high true-positive rates and low false-positive rates for nodule detection, the rib crossing areas and vessel spots produced camouflaging effect on nodule recognition. Besides, the sensitivity of the aforementioned detecting techniques was determined by the algorithmic parameter setting, e.g., less stringent the criteria, high sensitivity the detection. As a consequence, the detection algorithm would falsely classify the blood vessel, trachea, etc., as the pulmonary nodules of small size. Additional filtering method should be proposed to decrease the number of false positives. There are thus pressing needs for proposing novel pattern recognition approaches that can address the imposing challenges arising from false-positive elimination.

*2) Backpropagation Neural Networks*: To decrease the frequency of false-positive detection, several researches have been dedicated to finely filter the suspected nodules after the pre-scanning. One of the resorted filter techniques is the neural network, which is trained as a classifier to distinguish between the nodules and non-nodules among all of the suspected nodules detected by the pre-scan methods. It can be considered as a type of two-stage hybrid approaches. In the first stage, i.e., the pre-scanning stage, nodule candidates are initially extracted by using the feature extraction method, e.g., lung segmentation, morphological closing process, and rule-based thresholding algorithms; while in the second stage, the neural network based filters which take the sets of features from stage one as inputs, are to finely filter the suspicious nodule candidates so as to reduce the false positives.

After detecting the regions of suspected nodules with a sequential preliminary scanning procedure, i.e., thresholding evaluation and background subtraction for nodule enhancement (Giger et al. 1988), and sphere profile matching algorithm (Lo et al. 1993a), to discern the rib crossings and round vessels from the true nodules, Lo et al. (1993a) trained neural networks by backpropagation algorithm to classify the suspected nodules into three categories, i.e., non-nodule, nodule, and calcified nodules. A two-layer fully connected neural network was constructed where the input layer was composed of 1024 nodes and the input value for each node was the pixel value on the patch with 32x32 elements, meanwhile the hidden layer was composed of 100-200 nodes. Through the neural networks based classification, the false-positive nodules were detected so that an improved detection rate of true-positive nodules could be achieved. 30 chest radiographs consisting of 59 nodules and 175 non-nodules were used in this study and three experiments were performed (training set and test set are divided differently), the non-nodules were detected so that a reduced detection rate of false-positive could be achieved. Particularly, the AUC value of the experiment with the best performance reached 0.782.

So far, most researches were based on one consolidated data set. There were several studies based on multiple heterogeneous data sets. For instance, Torres et al. (2015) proposed a CAD system named M5L for nodule segmentation and detection. The M5L CAD employed ant colony based procedure to segment the lung nodules from other lung structures and made use of voxel-based feed-forward neural network of 1 hidden layer with 25 neurons for module classification based on 13 features, including spatial, intensity, and shape features. The training data set consisted of 69 scans from LIDC/IDRI (Lung Image Database Consortium – Image Database Resource Initiative) (Armato et al. 2010; Armato et al. 2011), 5 scans from ANODE09 (Van Ginneken et al. 2010), and 20 scans from ITALUNG-CT (Pegna et al. 2009) . In 949 scans from LIDC/IDRI, the M5L CAD achieved a sensitivity of about 80% at 8 false positives per scan. In the case of ANODE09, the M5L CAD performed a little worse. It also showed that the sensitivity was reduced for subtle nodules and ground glass opacities structures. Gupta et al. (2018) presented a two-stage approach for detection of multi–size pulmonary nodules in CT images from multiple heterogeneous data sets. The aim of the first stage was to initially extract different-sized nodule candidates by lung segmentation, morphological closing process, and rule-based thresholding algorithms. The purpose of the second stage was to reduce the false positives by a three-layered feed-forward neural network classifier which was composed of 515 input neurons, 49 hidden neurons, and 1 output neuron. A rich set of 515 features from cluster, texture, and voxel−based intensity features were utilized as the input. The proposed detection scheme was trained on the 899 CT scans from the LIDC-IDRI data set (Armato et al. 2010; Armato et al. 2011). The CAD system was tested on 70 scans from the AAPM–SPIE–LUNGx Dataset (Armato et al. 2015), 50 scans from the ELCAP (Early Lung Cancer Action Project) [http://www.via.cornell.edu/databases/lungdb.html] and 33 scans from PCF (Prevent Cancer Foundation) [http://www.via.cornell.edu/databases/crpf.html]. Detection results showed that a sensitivity of 85.6% at 8 false-positives per scan and 83.5% at 1 false-positive per scan on the LIDC-IDRI data set were achieved, respectively, and an average sensitivity of 68.4% at 8 false positives per scan was achieved for the three data sets.

In lung nodule diagnosis, it is well established that compared with the distant pixels, the close pixels' correlation is usually higher (Lo et al. 1993b), which suggests that the neighborhood information (e.g., interactions between nearby pixels) should be more emphasized than that between far distance pixels. However, in the aforementioned conventional backpropagation neural network model, the interactions between nearby pixels are assumed to be uniformly distributed. There thus are needs to adapt neural network model to accommodate the discriminative interactions between neighboring and non-neighboring pixels.

*3) Convolutional Neural Network*: Convolutional neural network (CNN) applies convolution operation to the input in the convolutional layers, which emulates the response of an individual neuron to visual stimuli (Fukushima et al. 1983). The main difference between the aforementioned conventional neural network and the CNN is that the conventional weights of the former network are independent while the kernel weights of the latter network are constrained by grouping. It is well established that the interactions between nearby pixels could be well taken into consideration in the framework of the CNN. Thus, in principle the region surrounding the suspected nodule could be scrutinized by CNN better than the conventional neural networks (Fukushima et al. 1983). Meanwhile, by observing the clinical radiologist's diagnosing procedure, it is found that the patches containing the suspected nodule(s) are usually examined in multiple different scales and angles. For instance, usually the clinical radiologist recognizes the disease patterns following a routine that whole pixel information is explored to detect promising regions containing suspected nodule(s), then the zones are subject to much more scrutiny to identify the presence of real nodule(s) (Lo et al. 1993b). The diagnosing process then starts all over again until all nodules are detected. The inherent properties of rotation invariance and shift invariance, which can be reserved within a small range of variance by the pooling layers of the CNN, are suitable for the recognition of disease patterns in medial image. Motivated by the aforementioned features of CNN, e.g., rotation and shift invariance, and description of complex interactions between nearby pixels, CNN has been adopted for recognition of pulmonary nodule. We want to make comment that the CNN before the deep learning era is classified to be the classical pulmonary nodule detection method in consideration that the CNN with only 3-4 layers is not as deep as the deep CNN in the deep learning era (LeCun et al. 2015).

Lo et al. (1993b) adopted the CNN for recognition of pulmonary nodule in gray scale images. In their work, a set of the feature extraction methods (Lo et al. 1993a) (e.g., boundary detection and sphere profile matching algorithms) were deployed for the initial detection of nodules. In this pre-scanning phase, the searching parameters were set in a highly sensitive level to identify all possible nodules like areas of 32x32 pixels. Before the extracted image patches containing possible nodules were input into the CNN, a background reduction method was employed to eliminate the background information as well as to enhance the contrast between possible nodules and its surroundings. Then, a CNN was structured as three-layer (two hidden layers), with 5x5 convolution kernel in each layer. The constructed CNN was trained through backpropagation algorithms without bias term and momentum term, and was designed to perform the differentiation of non-nodules from nodules. As kind of supervised leaning paradigm, to imitate the radiologists' diagnostic rating on the likelihood of nodules or non-nodules, a fuzzy output association method was constructed using Gaussian distribution and repulsive functions to turn the probability of suspicion in containing a disease into the information learnable by CNN on the output side. It is contended that the expected performance of CNN could be enhanced with the increasing number of training samples. In their work, they enlarged the sample size by taking the advantage of classification invariance property of matrix operations to enlarge the sample size. The original input patches (matrixes) went through the rotation operation while keeping the same output label. In specific, the original image block was manipulated with the four rotations of 0°, 90°, 180°, and 270° to obtain four blocks; the original image matrix was flipped over and manipulated with the above rotations again to obtain four additional patches. It should be remarked that the direct use of the rotation operations may not be suitable for those image patterns possessing circular symmetric nodules or lacking a fixed geometric pattern (e.g., calcification). In this work, the training image blocks were selected from 5 chest radiographs containing 42 nodules and 50 non-nodules, and the test image blocks were collected from 21 images containing 50 nodules and 258 non-nodules. The CNN achieved 80% true-positive detection rate with 2.6 false nodules per image. A number of false-positive detections were reduced and the superior performance of CNN suggested its potential use in clinic.

*4) Associative Learning Neural Networks*: Inspired by the features of learning and memory identified in neurobiological research (Olds et al. 1989), dynamically stable associative learning-based neural network was proposed to mimic the associative learning mechanism in the Hermissenda (Blackwell et al. 1992). In the associative learning model, when the Hermissenda experiences repeated pairs of conditioned stimulus followed by unconditioned stimulus, it learns the associative relationship between the two stimuli, and predicatively exhibits the reaction induced by unconditioned stimulus in response to conditioned stimulus alone. To incarnate the associative learning principles derived from the biological phenomena into designing of neural networks, the conditioned stimulus is denoted as the pattern to be classified, while the unconditioned stimulus as its desired classification. Both the conditioned stimulus and its associated unconditioned stimulus are inputted separately into one patch, which consists of another weight parameter to represent the frequency of this pair of combination. Each patch stores a single association between pattern to be classified (conditioned stimulus) and its desired classification (unconditioned stimulus). And then the whole set of patches collectively stores all of the patterns learned. The patches are connected to the output neuron. Based on the aforementioned network structure, the dynamically stable associative learning mainly involves patch creation, merging and deletion. The similarity between the input feature vector and the learned patterns are measured in terms of correlation index. Then, the pattern match classifier searches for the most similar pattern in the associated memory. If the maximum similarity is lower than a pre-defined value, the new feature pattern will be created and stored as a newly learned pattern in the memory. The learned pattern is then assigned to an associated class which is either a true or a false disease case. To sum up, the dynamically stable associative learning based neural network classifies by learning associations between patterns in a training set and their desired classification.

Lo et al. (1998) performed comparative studies of investigating the performances of several different neural network models in the recognition of lung nodules on chest radiographs. First, the dynamically stable associative learning model (Blackwell et al. 1992) was selected as the one of methods for comparison. Second, a CNN was structured as three-layer (two hidden layers) with 7x7 convolution kernel in each layer. Each hidden layer consists of 10 groups. The output layer has 10 nodes (2 categories) which were fully connected to the second hidden layer. Gaussian-like activation function was used between the input layer and the first hidden layer. The purpose of this activation function was to treat both low and high cumulated signals as false features that would eventually facilitate the classification process in the following layers. The sigmoid activation function was used for all layers other than the first hidden layer and backpropagation training was applied to adjust weights between any two adjacent layers. Their study implied that pattern classifiers such as dynamically stable associative learning model cannot function alone to analyze image blocks (patches) with substantial background structures since those background structures would contaminate the feature vector and lead to degradation of the machine observers' performance. For instance, in their test, the image blocks extracted from 62 chest radiographs were divided into two groups. The first group contained 91 true nodules and 247 non-nodules. The second group contained 95 nodules and 258 non-nodules. Two groups of images acted as the training set and test set alternately. The result showed that the CNN, which internally performed feature extraction and classification, achieved the superior performances (80% true-positive detection at 2.5 false-positive per image) to the pattern match neural network (60% true-positive detection at 7 false-positive per image).

*5) Massive Training Artificial Neural Networks*: The massive training artificial neural network (MTANN) is composed of a modified multilayer neural network where the layers are fully connected by the adjustable weights (Suzuki et al. 2005). The activation function of the units in the hidden layer is a sigmoid function, while that in the output layer a linear function. The MTANN can manipulate the image by inputting directly the pixel values on the region of interest. Suzuki et al. (2003) developed a multi-MTANN for reduction of false positives in lung nodule detection on CT images. The Multi-MTANN consists of multiple parallel arranged MTANNs. Each MTANN was trained by extracted patches of CT images containing nodule together with images containing different types of non-nodule. In their instance, five MTANNs were to distinguish nodules from various-sized vessels, and another four MTANNs to discern opacities. Then the outputs from nine MTANNs were integrated using the logical AND operation. The data set used consisted of 101 low-dose CT scans obtained from 71 different patients who volunteered to participate in a lung cancer screening program in Nagano, Japan, between 1996 and 1999. The results showed that the proposed classifier reached a sensitivity of 80.3% at an average 4.8 false-positive per scan.

Furthermore, Tajbakhsh and Suzuki (2017) compared two MTANNs with four deep convolutional neural networks (i.e., AlexNet, rd-CNN, LeNet architecture, and sh-CNN) in detection of lung nodules in CT images. They used a database of low-dose thoracic helical CT (LDCT) from 31 patients from 1996 to 1999 in Japan. Based on the LDCT database consisting of 38 scans with a total of 1057 slices of size 512 by 512, their experiments demonstrated that the performance of MTANNs was superior to the performances of the deep CNN. Specifically, MTANNs generated 100% sensitivity at 2.7 false positives per patient, while the best performing deep CNN achieved the same level of sensitivity with 22.7 false positives per patient.

*6) K-Nearest Neighbors Classifier*: The *k*-nearest neighbors classifier is a non-parametric procedure according to which an unclassified case is assigned to the class represented by a majority of its *k* nearest neighbors from the training set (Cover and Hart 1967). Murphy et al. (2009) applied two successive *k*-nearest neighbors classifiers to reduce the false-positive nodules in CT images. In particular, after down-sampling of the image and segmentation of the lung volume (Hu et al. 2001; Sluimer et al. 2005), the shape index (SI) and curvedness (CV) features (Koenderink 1990) were adopted to detect suspicious nodule candidates. Once the SI and CV values for the image were known, seedpoints were established based on narrow threshold ranges for SI and CV values. Then the seedpoints were expanded to form clusters of voxels of interest by hysteresis thresholding (Canny 1986) and the clusters close to each other were merged to associate the separate portions of surface. Then two consecutive *k*-nearest neighbors classifiers were to perform the false-positive reduction, where the feature selection was carried out by Sequential Forward Floating Selection method (Pudil et al. 1994). Three databases were built based on the Dutch-Belgian Randomized Lung Cancer Screening Trial (NELSON) data (Heuvelmans et al. 2017; Horeweg et al. 2014). The first database was with 1535 scans (2894 nodules), the second and the third with 1121 scans each (3451 and 1528 nodules respectively). Evaluated on three databases, their approach achieved a sensitivity of 80% with an average 4.2 false positives per scan in 813 scans of first database, and a sensitivity of 72.4% / 77.7% with an average 4.0 / 4.2 false positives per scan in 541 scans of second/third database.

*7) Support Vector Machines*: Support vector machines are supervised learning algorithms, the main idea

behind which is to find the hyperplane maximizing the separation between classes (Rosales-Perez et al. 2018; Schölkopf and Smola 2001; Vapnik 1995). There has been interest in adopting support vector machines to pulmonary nodule diagnosis problems. Tan et al. (2011) applied a hybrid feature selection and classification methodology for detection of lung nodules in the LIDC database (Armato et al. 2010; Armato et al. 2011). In the detection stage of their approach, a nodule segmentation method based on nodule and vessel enhancement filters and a computed divergence feature was used to locate the centers of the nodules' clusters. In the classification stage, invariant features were used to differentiate between real nodules and some forms of blood vessels. Three feature-selective classifiers were presented, i.e., support vector machines, neural networks trained by genetic algorithms, and fixed-topology neural networks trained by Levenberg-Marquardt backpropagation algorithm. 235 randomly selected cases from LIDC database were used to train the classifiers, while the remaining 125 cases were for test. It was observed that all three classifiers performed comparably well. The neural networks and the fixed-topology neural networks achieved the highest sensitivity of 87.5% with an average of 4.0 false-positives per scan on 80 annotated nodules with agreement level 4 (the list of all the nodules marked by at least 4 of the four radiologists), while the support vector machines reached a sensitivity of 83.8% with an average of 4.0 false-positives per scan on the same agreement level of 4.

Setio et al. (2015) proposed a radial basis support vector machine classifier to identify large pulmonary nodules larger than 10 mm in thoracic CT scans. The proposed detection approach first employed a three-dimensional lung segmentation algorithm with the aim not to rule out the large nodules attached to the pleural wall. After masking out structures outside the pleural space, each lung image was resampled to an isotropic image with a resolution of 1.0 mm using Gaussian filter. Next, thresholding and morphological operations were deployed to obtain nodule candidates. A set of 24 indexes based on shape, intensity, blobness, and spatial context was calculated and extracted as the features. The SVM was used to classify nodule candidates. Evaluated on the 888 CT scans from the LIDC–IDRI, the proposed algorithm reached sensitivities of 98.3% and 94.1% on large nodules at an average of 4.0 and 1.0 false-positives per scan, respectively.

Bergtholdt et al. (2016) proposed a cascaded support vector machine classifier for addressing the detection of lung nodules in CT images. In the proposed cascaded SVM, two classification tasks were sequentially performed. In the pre-selection stage, a large set of suspicious pulmonary nodule candidates was scrutinized to reduce the probability that a true nodule was falsely rejected. Tested on the 1018 cases of LIDC-IDRI database (Armato et al. 2010; Armato et al. 2011), the proposed scheme achieved a sensitivity of 85.9% with 2.5 false-positives per volume.

Teramoto and Fujita (2018) proposed an automated detection scheme for nodule detection in both PET and CT images from East Nagoya Imaging Diagnosis Center of Nagoya in Japan. In CT images, the nodules were enhanced by cylindrical nodule-enhancement filter. Later, high-uptake regions detected by PET images were merged with the region detected by CT image. Based on the features extracted from both CT and PET images, rule-based classifier and three support vector machines eliminated false positives from the initial candidates. Evaluated on the 100 cases of PET/CT images, the sensitivity of the hybrid approach was 16% higher than the independent detection systems using only CT images, and the hybrid approach reached a sensitivity of 83% with 5 false-positives per case.

*8) Linear Discriminant Classifier and GentleBoost Classifiers*: Jacobs et al. (2014) presented a two-stage classifier for classifying subsolid nodules in CT images, as well as novel image-level context features describing the relationship of an area of ground glass opacity to its surroundings, e.g., lung, airways, vessels and other nodule candidates. Then, a rich set of 128 features was defined based on the proposed context features and the conventionally used intensity, shape, texture features (Kim et al. 2005; Tao et al. 2009; Ye et al. 2009; Zhou et al. 2006). To make accurate classification of subsolid nodules, first five features were selected to perform a first-round reduction of false-positives by Linear Discriminant Classifier so as to remove as many false-positive candidates as possible. Then, the remaining 119 features were computed and a combination of 10 GentleBoost classifiers was used to classify all candidates into the two classes. Experiments showed that image-level context features significantly improved the classification performance. Performance analysis on data set (209 scans for training and 109 scans for testing) from NELSON trial [a multi-center lung cancer screening trial organized in the Netherlands and Belgium (van Klaveren et al. 2009)]showed that the proposed two-stage classifier reached a sensitivity of 80% at an average 1.0 false positive per scan.

### *B. Deep Learning-Aided Decision Support for Pulmonary Nodules Detection*

The classical detection methods are ad hoc analysis schemes, which are confronted with very convoluted and tedious parameter tuning issues. From above analysis, it is established that the classical detection scheme first relies on conventional image processing techniques as pre-scan to detect the suspected nodules. The sensitivity of the techniques was determined by the algorithmic parameter setting, which leads to the changes of the amount of suspected nodules. Then the computer-assisted diagnosis scheme is assisted with the pattern recognition techniques to distinguish between the true nodules and non-nodules. However, the performances of the latter discern procedure depend heavily on the performance of the pre-scan step.

Deep learning uses cascaded multiple layers of nonlinear processing units to extract and learn multiple levels of features. To date, deep learning techniques have emerged as promising decision supporting approaches to automatically analyze medical images for different clinical diagnosing purposes (Bibault et al. 2016; Esteva et al. 2017; Greenspan et al. 2016; Gulshan et al. 2016; LeCun et al. 2015; Litjens et al. 2017; Miotto et al. 2017; Ravì et al. 2017; Schmidhuber 2015; Zhou et al. 2017), also demonstrate remarkable and significant progress in pulmonary nodules diagnosis, e.g., feature extraction, nodule detection, false-positive reduction. It is well established that the deep learning techniques seamlessly integrate exploitation feature and tuning of performance so as to simplify the ad hoc analysis pipeline of conventional computer-assisted diagnosis. In this regard, several studies were dedicated to pulmonary nodule detection with the assistance of deep learning techniques which are summarized in Table 1.

*1) Two-dimensional Convolutional Neural Networks*: Setio et al. (2016) addressed the false-positive reduction by proposing a novel two-dimensional multi-view deep convolutional neural network, where the inputs were fed with 2D patches extracted from differently oriented planes. Their approach reached sensitivities of 85.4% and 90.1% at 1 and 4 false-positives per scan, respectively, on 888 scans of LIDC-IDRI data set.

Jiang et al. (2018) proposed lung nodule detection scheme based on a four-channel deep CNN. First, a slope analysis method was designed to mend the distorted lung contours caused by juxta-pleural nodule, so that the lung parenchyma can be extracted from CT image perfectly. Second, Frangi filters were used to eliminate the vascular structure in the lung to enhance multi-group patches cropped from the lung images. After that, the threshold processing method was performed on the vessel-eliminated image. Then multi-group patches were built as follow: one group consisted of nodule patches cropped from the original lung images, and another group consisted of binary images (contained only background and suspicious regions) cropped from the vessel-eliminated images. To further improve the detection performance, these nodules were divided into four levels from the smaller to the larger sizes. Finally, through combining two groups of image data, a four-channel CNN model was designed to learn radiologists' knowledge for detecting nodules of four levels. The experiment was carried on the LIDC/IDRI data set with 25,723 nodules (with diameter greater than 3mm) in 1006 cases. The results indicated that the multi-group patch-based deep learning network achieved a sensitivity of 80% with 4.7 false positives per scan and the sensitivity of 94% with 15.1 false positives per scan.

Nam et al. (2019) designed a deep learning–based automatic detection algorithm (DLAD) for detecting malignant lung nodules in chest radiographs. Structured by 25 layers and 8 residual connections based on deep residual learning (He et al. 2016), DLAD used the pixel intensity of chest radiography as the input, and the disease classification and lesion locations as two outputs. DLAD was trained in a semi-supervised learning manner using all image-level annotations (normal or malignant nodules), but only part of the pixel-level annotations (location of malignant nodules). The outputs of 3 networks learned on the same data but with different hyper-parameters were averaged for the final prediction. 43,292 chest radiographs collected from Seoul National University Hospital (normal-to-malignant ratio, 34,067:9225) were divided into three parts for the purpose of training, tuning and validation (normal-to-malignant ratio, 33,467:8625; 300:300; 300:300). Four external data sets were prepared for validation (normal-to-malignant ratio, 62:119 from Seoul National University Hospital; 59:123 from Boramae Hospital; 70:111 from National Cancer Center of Korea; 60:89 from UCSF Medical Center). On the five validation data sets, DLAD attained AUC of 0.92-0.99 and JAFROC FOM of 0.831-0.924 in radiograph classification and malignant nodule detection tasks, respectively. For radiograph classification, DLAD reached sensitivities of 79.0%, 91.1%, 71.2%, and 87.6% with specificities of 95%, 98%, 100%, and 93% on four external data sets. For malignant nodule detection, DLAD reached sensitivities of 69.9%, 82.0%, 69.6%, and 75.0% at 0.34, 0.30, 0.02, and 0.25 false positives per image. 18 physicians with different qualifications participated in an observer performance test. The results showed that the DLAD was significantly ($P < 0.05$) superior to 11 physicians in radiographs classification and 15 physicians in nodules detection. In addition, most physicians assisted with DLAD (14 in the classification task and 15 in the detection task) significantly ($P < 0.05$) improved their performances. However, DLAD had limitations in detecting small (< 1 cm) or less conspicuous nodules, since DLAD was trained with the labeled data which were affected by human perception. Moreover, DLAD did not detect any retrophrenic nodules since it was undertrained to focus on the retrophrenic area.

*2) Three-dimensional Convolutional Neural Network*: To take full advantage of 3D spatial contextual information of pulmonary nodules, Dou et al. (2017) proposed sophisticated three-dimensional CNNs for false positives reduction of pulmonary nodule in volumetric CT scans, where the input of each of the three different 3D CNN was fed with image with different receptive field so that multiple levels of contextual information surrounding the suspicious nodule could be encoded. Then the prediction probabilities from the 3D CNNs for given candidates were fused with weighted linear combination. Evaluated on 888 CT scans from the LIDC data set, sensitivity of all three designs exceeded 90% with 8.0 false positives per scan, and the fusion model achieved a sensitivity of 92.2% under the same false-positive rate.

*3) Convolutional Autoencoder Neural Network*: To properly train the deep learning model, a large data set of labeled data is usually required. However, due to the issues arising from privacy protection and labor-intensive labeling tasks, it is great challenge to set up large labeled data set for assistance of deep network functioning. In the literature it can be found that the majority of deep learning researches for detection and classification tasks are based on supervised learning scheme. However, to tackle the critical issues in shortage of data, it is likely that the possible steps to this direction could be semi-supervised or even unsupervised learning. In the direction, Chen et al. (2018) introduced the convolutional autoencoder neural network (CANN) to perform unsupervised feature learning for classification between nodules and non-nodules. In the proposed framework, the image patches were used for unsupervised feature learning, with a small amount of labeled data for supervised parameters' fine-tuning. Once the feature representations have been extracted, the CT images were input into the CANN, and the patch images were extracted for classification between nodules and non-nodules. They compiled a data set from a hospital in China, including about 4500 patients' lung CT images from 2012 to 2015. Comparison was based on the data set with 3662 patches (1754 nodule patches and 1908 normal patches). Experimental results showed that the proposed CANN could accurately extract the pulmonary nodule's features (accuracy of 95% and AUC of 0.98), and achieve fast labeling while avoiding the intensive laboring on labeling via the hand-crafted features.

*4) Ensemble Method*: Teramoto et al. (2016) reduced the false positives of solitary pulmonary nodules by proposing an ensemble method. In their study, both PET and CT images were used which provided metabolic and anatomical information, respectively. This ensemble method took advantage of not only the specifically defined features about PET and CT images by the physician's expertise, but also the automatically extracted features by the confirmation of the CNN. First, the CT images were enhanced by an active contour filter to detect possible regions; the PET images were binarized by a predetermined threshold to detect high-uptake regions. Both suspicious regions from PET/CT images were combined using logical OR in the pixel-by-pixel level and treated as the initial candidate regions. Second, two feature extraction methods were developed, i.e., the shape and metabolic feature analysis and the CNN-based feature extraction. In particular, a set of 18 features from CT images was selected, including sectional areas, volume, surface area, contour pixels, compactness, convergence, max, center, and standard deviation. A set of 8 metabolic features from PET images was obtained, including the standardized uptake value (SUV), the maximum and mean values of SUV, sectional areas, volume, and surface area. Third, a two-step classifier was developed in which rule-based classifier worked in the first stage, and two SVMs worked in the second stage. One SVM took 22 features (all features by CT, three SUV features by PET, and features extracted by CNN) as the input, while another SVM took all features as input. Evaluation based on 104 PET/CT images showed that the ensemble method achieved a sensitivity of 90.1% with 4.9 false positives per case.

### *C. Comparisons and Contests for Detection and False Positive Reduction Algorithms*

It has been witnessed from aforementioned investigation that to deal with the grand challenges on how to accurately detect the pulmonary nodules with high sensitivity at low false-positives rates, consistent endeavors have been exerted to evolve the computer-assisted nodules detection techniques from the complicated ad hoc analysis pipeline of conventional approaches to the simplified seamlessly integrated deep learning techniques. The development course of computer-assisted nodules detection techniques over 3 decades is depicted in Fig. 4. To further demonstrate the performance differences, we compare the classical nodule detection methods and deep learning aided methods in terms of sensitivity, false positive rate, and AUC, which is showed in Table 2.

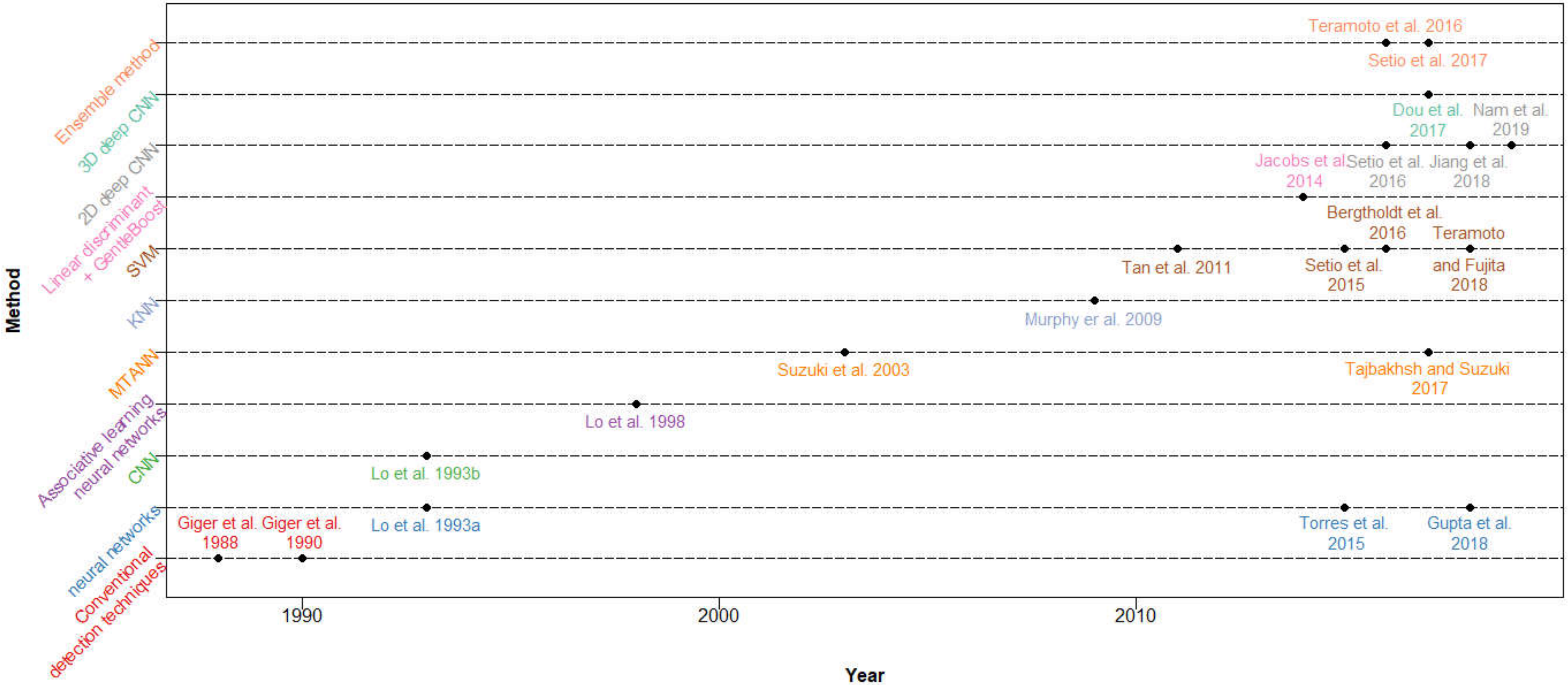

**Fig. 4** Development course of computer-assisted nodules detection techniques from complicated ad hoc analysis pipeline of conventional approaches to the simplified seamlessly integrated deep learning techniques over 3 decades

**Table 2** Comparison between classical nodule detection method and deep learning-aided method

| ***Method*** | | ***Author*** | ***Sensitivity*** | ***False positive rate*** | ***AUC*** | ***Data set*** |
|---|---|---|---|---|---|---|
| Classical Detection Methods | Conventional detection techniques | Giger et al. (1988) | 80% | 2.7 per image | -[b] | Authors' compiled data set |
| | | Giger et al. (1990) | 57.6% | 3.7 per image | - | Authors' compiled data set |
| | Neural network | Lo et al. (1993a) | - | - | 0.782 | Authors' compiled data set |
| | | Torres et al. (2015) | 80% | 8 per scan | - | LIDC-IDRI, ITALUNG-CT, and ANODE09 |
| | | Gupta et al. (2018) | 85.6% | 8 per scan | 0.957 | LIDC-IDRI |
| | | | 66.3% | 8 per scan | 0.831 | AAPM-SPIE-LUNGx |
| | | | 70.4% | 8 per scan | 0.847 | ELCAP |
| | | | 68.9 | 8 per scan | 0.804 | PCF |
| | CNN | Lo et al. (1993b) | 80% | 2.6 per image | 0.88 | Authors' compiled data set |
| | Associative learning neural networks | Lo et al. (1998) | 80% | 2.5 per image | - | Authors' compiled data set |
| | MTANN | Suzuki et al. (2003) | 80.3% | 4.8 per scan | - | Authors' compiled data set |
| | | Tajbakhsh and Suzuki[a] (2017) | 100% | 2.7 per scan | 0.8806 | Authors' compiled data set |
| | KNN | Murphy et al. (2009) | 80.0% | 4.2 per scan | - | NELSON |
| | SVM | Tan et al. (2011) | 83.8% | 4.0 per scan | - | LIDC |
| | | Setio et al. (2015) | 94.1% / 98.3% | 1.0 / 4.0 per scan | - | LIDC-IDRI |
| | | Bergtholdt et al. (2016) | 85.9% | 2.5 per scan | - | LIDC-IDRI |
| | | Teramoto and Fujita (2018) | 83.5% | 5 per scan | - | East Nagoya Imaging Diagnosis Center |
| | Linear discriminant classifier + GentleBoost classifiers | Jacobs et al. (2014) | 80% | 1.0 per scan | - | NELSON |
| Deep learning aided method | 2D deep CNN | Setio et al. (2016) | 85.4% / 90.1% | 1 / 4 per scan | - | LIDC-IDRI |
| | | Jiang et al. (2018) | 80.06% / 94% | 4.7 / 15.1 per scan | - | LIDC-IDRI |
| | | Tajbakhsh and Suzuki [a] (2017) | 100% | 22.7 per scan | 0.7755 | Authors' compiled data set |
| | | Nam et al. (2019) | 69.9%, 82.0%, 69.6%, and 75.0% | 0.34, 0.30, 0.02, and 0.25 per image | 0.885, 0.924, 0.831, and 0.880 | Authors' compiled data set (evaluated on four external validation data sets) |
| | 3D deep CNN | Dou et al. (2017) | 92.2% | 8.0 per scan | - | LIDC |
| | Ensemble method | Teramoto et al. (2016) | 90.1% | 4.9 per scan | - | Authors' compiled data set |
| | | Setio et al. (2017) | 96.9% / 98.2% | 1.0 / 4.0 per scan | - | LIDC-IDRI |

[a] Tajbakhsh and Suzuki compared performance of MTANN and CNN, which were listed in two entries in the table

[b] The sign "-" means missing data

To provide fair comparisons of different detection and false positive reduction algorithms, Setio et al. (2017) set up the LUng Nodule Analysis 2016 (LUNA16) challenge based on the LIDC-IDRI data set of 888 chest CT scans (Armato et al. 2010; Armato et al. 2011). Seven systems have been applied to the complete nodule detection track and five systems have been applied to the false positive reduction track. They investigated the impact of combining individual systems on the detection performances. It was observed that the leading solutions employed CNN and used the provided set of nodule candidates. The combination of these solutions achieved an excellent sensitivity of 96.9% at 1.0 false-positive per scan. Moreover, the LIDC-IDRI reference standard has been updated by identifying nodules that were missed in the original LIDC-IDRI annotation. In 2017, 1972 teams participated in the Data Science Bowl (DSB) [https://www.kaggle.com/c/data-science-bowl-2017/], presented by Booz Allen Hamilton and Kaggle. The focus of the DSB2017 was to develop lung cancer detection algorithms for CT images. A total of 1397, 198, and 506 cases were used for training, validation, and testing. The ground truth labels were determined by pathology diagnosis. Moreover, the LUNGx Challenge (Armato et al. 2016; Armato et al. 2015), supported by the International Society for Optics and Photonics (SPIE), American Association of Physicists in Medicine (AAPM) and the National Cancer Institute (NCI), was conducted for computerized classification of lung nodules as benign or malignant in CT scans. 11 methods were applied to classify 73 nodules (37 benign and 36 malignant). The AUC value for these methods ranged from 0.50 to 0.68. Three methods performed statistically better than random guessing.

## III. Computer-assisted Benign-malignant Classification of Pulmonary Nodule

There is consistent endeavor to deal with the grand challenges on how to automatically and precisely differentiate between benign and malignant nodules on different image formats to support decision making. Differentiation of the malignant/benign nature of a nodule traditionally involves two consecutive tasks, i.e., image processing and classification. The purpose of image processing is to extract the features which serve as the input for the latter classification task. Therefore, extraction of reliable features and combination of proper features are two primary concerns for nodule benign/malignant classification. Based on the feature extraction and combination schemes, we roughly divide them into three categories—classical classification techniques, classical classification enhanced by deep learning based feature learning, and deep learning-aided pulmonary nodules classification techniques. The classical methods heavily depend on image processing for well-defined explicit features of a disease pattern, and then recognize/determine the benign/malignant lesion/tumor by maximizing the similarity between the profiles of selected features and the suspected nodules. Since the performances of the classical discerning procedure depend heavily on the set of selected features, to alleviate the negative effect from inappropriate or incomplete feature set, the classical techniques are enhanced by deep learning which learns the features from data and automatically extracts features in place of the features defined by the expertise. Distinct from the ad hoc analysis pipeline of classical methods and their enhanced versions by deep learning, the deep learning aided methods seamlessly integrate the feature computing, selection, and composition for classification. The illustrative framework of computer-assisted pulmonary nodules classification is depicted in Fig. 5.

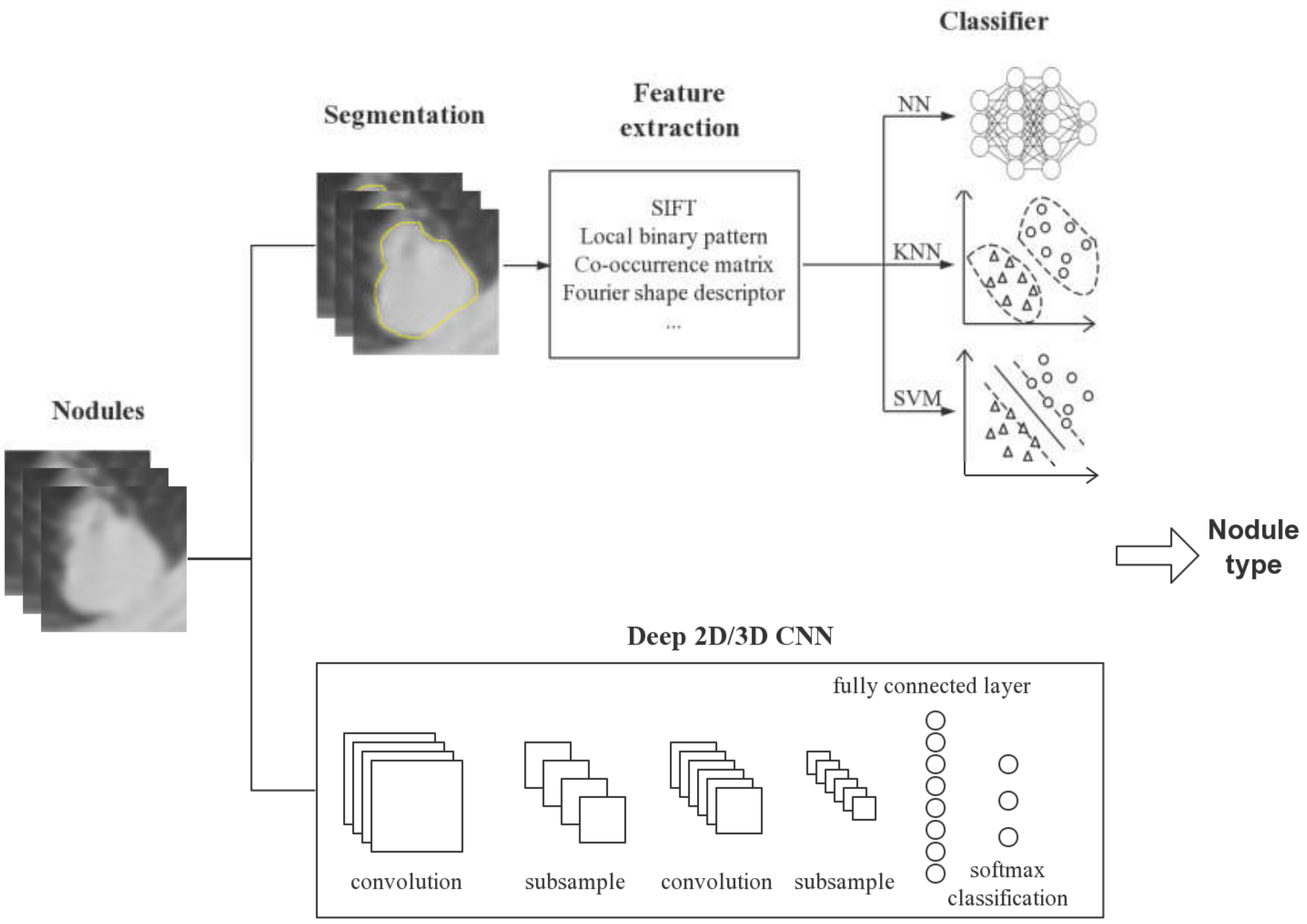

**Fig. 5** Illustrative framework of computer-assistant pulmonary nodules classification

In this section, we analyze the studies dedicated to pulmonary nodule classification with the three main stream techniques, the classical techniques, classical classification enhanced by deep learning-based feature learning, and deep learning techniques, which are summarized in Table 3.

**Table 3** Benign-malignant classification techniques for pulmonary nodules

| *Authors* | *Year* | *Methods* | *Learning Methodology* | *Effects* | *Data set Size* | *Data Dimension* | *Image Format/data source* |
|---|---|---|---|---|---|---|---|
| Suzuki et al. (2005) | 2005 | (1) Feature computation: multiple massive training artificial neural networks<br>(2) Classifier: ANN | Supervised learning | AUC of 0.882;<br>Malignant nodules: sensitivity of 100% with specificity of 48.4% | 76 primary lung cancers in 73 patients and 413 benign nodules in 342 patients | 2D | Chest CT |
| Hua et al.[a] (2015) | 2015 | (1) Feature extraction: scale invariant feature transform (SIFT) and local binary pattern (LBP)<br>(2) Classifier: *k*-nearest neighbors classifier | Supervised learning | Malignant nodules: sensitivity of 75.6% with specificity of 66.8% | 2545 nodules from 1010 scans | 2D | Chest CT / LIDC |
| Hua et al.[a] (2015) | 2015 | (1) Feature computation: fractional Brownian motion model to estimate the Hurst coefficients<br>(2) Classifier: support vector machine | Supervised learning | Malignant nodules: sensitivity of 50.2% with specificity of 57.2% | 2545 nodules from 1010 scans | 2D | Chest CT / LIDC |
| Hua et al.[a] (2015) | 2015 | Classifier: deep belief network (DBN) | Unsupervised learning | Malignant nodules: sensitivity of 73.4% with specificity of 82.2% | 2545 nodules from 1010 scans | 2D | Chest CT / LIDC |
| Hua et al.[a] (2015) | 2015 | Classifier: CNN | Supervised learning | Malignant nodules: sensitivity of 73.3% with specificity of 78.7% | 2545 nodules from 1010 scans | 2D | Chest CT / LIDC |
| Tajbakhsh and Suzuki (2017) | 2017 | Classifiers: massive-training artificial neural networks (MTANNs), AlexNet, rd-CNN, LeNet, and sh-CNN | Supervised learning | AUC of 0.8806 by MTANNs;<br>Best AUC of 0.7755 by CNN | 76 primary lung cancers in 73 patients and 413 benign nodules in 342 patients | 2D | Chest CT |
| Tu et al. (2017)[e] | 2017 | Classifier: Two-dimensional CNN | Supervised learning | Accuracy of 91.7 ± 1.3 | 190 nodules for each of three classes (solid, | 2D | Chest CT / LIDC |
| | | | | | part-solid and non-solid) from 1,010 scans | | |
| Shen et al. (2017) | 2017 | Classifier: Multi-crop CNN | Supervised learning | Accuracy of 87.14%; AUC of 0.93 | 880 benign nodules, 495 malignant nodules, and 1,243 uncertain nodules | 2D | Chest CT / LIDC-IDRI |
| Sun et al. (2017) | 2017 | Classifier: LeCun CNN, deep belief network, and stacked denoising autoencoder (SDAE) | Supervised learning | AUC of 0.899 ± 0.018 by CNN | 1018 scans (41,372 benign nodules and 47,576 malignant nodules after data augmentation) | 2D | Chest CT / LIDC |
| Ciompi et al. (2017) [e] | 2017 | Classifier: multi-stream multi-scale CNN | Supervised learning | Accuracy of 79.5% for six nodule types | 1805 nodules (S / C / Pa / N / Pe / Sp: 926 / 311 / 84 / 202 / 243 / 39) from the MILD + 639 nodules (S / C / Pa / N / Pe / Sp: 382 / 58 / 37 / 87 / 48 / 27) from the DLCST [c] | 2D | Chest CT / MILD and DLCST |
| Liu et al. (2017) | 2017 | Classifier: three-dimensional CNN | Supervised learning | AUC of 0.732 by 3D CNN; AUC of 0.780 by ensemble models with 3D CNN | 163 benign nodules + 163 malignant nodules | 3D | Chest CT / NLST and ELCAP |
| Nishio et al. (2018a) | 2018 | 1) Feature computation: local binary pattern (LBP) in 3D image<br>(2) Classifier: SVM and XGBoost, parameter optimized by Tree Parzen Estimator (TPE) and random search | Supervised learning | AUC of 0.850 and accuracy of 79.7% by SVM + TPE;<br>AUC of 0.896 and accuracy of 82.0% by XGBoost + TPE | 36 malignant nodules and 37 benign nodules from LUNGx challenge; 26 malignant nodules from NSCLC | 3D | Chest CT / LUNGx challenge and NSCLC |
| Nishio et al. (2018b) | 2018 | Classifier: deep CNN modified by VGG-16 CNN | Supervised learning and | Validation accuracy of 68.0% for three nodule types | 412 benign nodules + 571 primary lung | 2D | Chest CT |
| | | | transfer learning | | cancers + 253 metastatic lung cancers | | |
| Xie et al. (2018) | 2018 | (1) Feature extraction: co-occurrence matrix, Fourier shape descriptor and deep CNN<br>(2) Classifier: AdaBoosted back propagation neural network | Supervised learning | AUCs of 0.9665, 0.9445 and 0.8124 for three data sets, respectively;<br>Accuracies of 89.53%, 87.74%, 71.93% for three data sets, respectively;<br>Malignant nodules: Sensitivities of 84.19%, 81.11%, and 59.22% with specificity of 92.02%, 89.67%, and 84.85% for three data sets, respectively | The first data set contains 1324 benign nodules and 648 malignant nodules, the second contains 2021 benign nodules and 648 malignant nodules, the third contains 1324 benign nodules and 1345 malignant nodules. | 2D | Chest CT / LIDC-IDRI |
| Yuan et al. (2018) [e] | 2018 | (1) Feature extraction: multi-view multi-scale CNN and Fisher vector encodings<br>(2) Classifier: multi-class support vector machine | Supervised learning | Accuracies of 93.1% and 93.9% for two data sets, respectively | 1738 nodules (W / P / V / G / J: 905 / 329 / 219 / 82 / 203) and 1000 non-nodules from 744 scans on the LIDC-IDRI + 421 nodules (W / P / V / G / J: 92 / 155 / 49 / 19 / 106) from 46 cases on the ELCAP [d] | 2D | Chest CT / LIDC-IDRI, ELCAP |
| Liu et al. (2018) [e] | 2018 | Classifier: multi-view multi-scale CNNs | Supervised learning | Accuracies of 92.1% and 90.3% on the two data sets, respectively | 1738 nodules (W / P / V / G / J: 905 / 329 / 219 / 82 / 203) and 1000 non-nodules from 744 scans on the LIDC-IDRI | 2D | Chest CT / LIDC-IDRI and ELCAP |
| | | | | | + 421 nodules (W / P / V / G / J: 92 / 155 / 49 / 19 / 106) from 46 cases on the ELCAP [d] | | |
| Hussein et al. [b] (2018) | 2018 | (1) Feature extraction: 3D CNN<br>(2) Classifier: graph regularized sparse multi-task learning | Supervised learning and transfer learning | Accuracy of 91.26% | 635 benign nodules + 509 malignant nodules | 3D | Chest CT / LIDC-IDRI |
| Hussein et al. [b] (2018) | 2018 | (1) Initial label set: clustering<br>(2) Classifier: proportion-SVM | Unsupervised learning | Accuracy of 78.06%; Malignant nodules: sensitivity of 77.85%, and specificity of 78.28% | 635 benign nodules + 509 malignant nodules | 2D | Chest CT / LIDC-IDRI |

[a] Hua et al. compared performances of SIFT plus LBP, SVM, DBN, and CNN, which were listed in four entries in the table

[b] Hussein et al. compared performances of 3D CNN and proportion-SVM , which were listed in two entries in the table

[c] S = Solid; C = Calcified; Pa = Part-solid; N = Non-solid; Pe = Perifissural; Sp = Spiculate

[d] W = Well-circumscribed; P = Pleural-tail; V = Vascularized; G = Ground glass optic; J = Juxta-pleural

[e] These paper classified nodule types rather than classifying malignant-benign nodules

*A. Classical Classification Techniques*

*1) K-nearest Neighbors classifier*: Hua et al. (2015) implemented the *k*-nearest neighbors classifier for benign-malignant classification in CT images as baseline comparison algorithm with deep belief network, and deep CNN. In the *k*-nearest neighbors classifier, first, geometric descriptors, i.e., scale invariant feature transform (SIFT) and local binary pattern (LBP), were adopted as quantitative features profiling the two-dimensional region of interest. Then, dimension reduction schemes were introduced to shrink the size of features while preserve the discriminative power. Finally, the shrunk set of features was fed to the *k*-nearest neighbors classifier for classification task. Evaluated on the subset of LIDC date set in which the nodules with diameters larger than 3 mm were selected (i.e., 2545 nodules from 1010 cases), the results showed that the *k*-nearest neighbors classifier reached a sensitivity of 75.6% and a specificity of 66.8%.

*2) Support Vector Machine*: Hua et al. (2015) also implemented the SVM as baseline for illustration of the effectiveness of the deep learning techniques with regard to the nodule classification problem. Specifically, the fractional Brownian motion model was adopted to estimate the Hurst coefficient, which was shown to be linearly correlated with the fractal dimension, at a defined neighborhood. Five coefficients were computed with respect to the neighborhood radius of 3, 5, 7, 9, and 11, respectively, as the feature vector. SVM was utilized to identify the malignant or benign nature of the nodule based on the computed feature vector. A total of 2545 annotated nodules in 1010 cases were selected from the LIDC date set. The results showed that the SVM enhanced by fractal analysis reached a sensitivity of 50.2% and a specificity of 57.2%.

Nishio et al. (2018a) explored Bayesian optimization (Tree Parzen Estimation) for optimizing parameters of both SVM and gradient tree boosting (XGBoost). Feature vectors were calculated by LBP from cropped CT images and used to train SVM/XGBoost with the nodules' corresponding classification labels. Parameters in SVM and XGBoost were then optimized with both random search method and Bayesian optimization. The results showed that both SVM and XGBoost optimized by Bayesian optimization achieved better performance, comparing to random search method.

*3) Massive Training Artificial Neural Networks*: Suzuki et al. (2005) developed a multiple-massive training artificial neural networks (multiple MTANNs) to distinguish malignant nodules from benign nodules in CT images. Each MTANN was trained by input images and teaching images with distribution estimation of likelihood of being a malignant nodule. Then, the outputs of the multiple MTANNs were combined by an integration neural network to discern the benign nodules from malignant ones. Based on a consolidated database consisting of 76 primary lung cancers in 73 patients and 413 benign nodules in 342 patients in Japan, the multiple-MTANNs achieved an AUC of 0.882, and can identify all malignant nodules with 48% benign recognition.

Furthermore, Tajbakhsh and Suzuki (2017) compared two MTANNs and four state-of-the-art CNN (AlexNet, rd-CNN, LeNet architecture, and sh-CNN) in distinction between benign and malignant lung nodules in CT images. They used a date set consisting of 76 histopathologically confirmed lung cancers in 73 patients and 413 benign nodules in 342 patients. The nodule size ranged from 3 mm to 29 mm. Their experiments demonstrated that the performances of MTANNs were higher than those of CNNs. The MTANNs yielded AUC of 0.8806, while the best performing CNN achieved an AUC of 0.7755.

*B. Classical Classification Enhanced by Deep Learning-Based Feature Learning*

Instead of using the handcrafted features, a few of classical nodule classification algorithms have been enhanced and fed by automatically generated features from the deep learning. In particular, the outputs from layers of the network are extracted and considered as features which are used to train a separate pattern classifier. It is expected that fusing of CNN-based features and handcrafted features from expertise can yield significantly improved performances.

*1) Backpropagation Neural Networks Plus Deep CNN Feature Learning*: Xie et al. (2018) made differentiation between the malignant and benign nodules by building an ensemble classifier in which the classification decisions from three AdaBoosted back propagation neural networks (Adaboosted BPNN) were weighted summed. Each AdaBoosted BPNN was trained by one of feature descriptors, i.e., the gray level co-occurrence matrix (GLCM)-based texture descriptor, a Fourier shape descriptor and an eight-layer deep convolutional neural network based on LeNet-5 (Lecun et al. 1998). The accuracy on the validation set of each Adaboosted BPNN served as the weight of decision in the ensemble classifier. Evaluated on the three data sets based on LIDC-IDRI (1018 scans), in which the nodules with a composite malignancy rate 3 were discarded, regarded as benign or regarded as malignant, the proposed algorithm achieved an AUC of 96.65%, 94.45% and 81.24%, respectively, higher than the AUC obtained by the LeNet-5 feature, GLCM-based texture descriptor and Fourier shape descriptor, respectively.

*2) Multi-class Support Vector Machine Plus Deep CNN Feature Learning*: Yuan et al. (2018) proposed a classification model for different lung nodule types including well-circumscribed, vascularized, juxta-pleural, pleural-tail, ground glass optical and non-nodule. First they divided nodule volumes by icosahedron pattern to capture both nodules and their surrounding anatomical structures. With the sampling results and intensity analysis, a threshold based method was applied to approximate nodules radii, followed by high frequency content analysis view sorting. Based re-sampled views, they trained multi-view multi-scale CNNs to describe nodule statistical features and calculated the Fisher vectors by dense SIFT method to represent geometrical features. Finally, multiple kernel learning was utilized to fuse the features and the nodule types were classified by support vector machine according to hybrid features. They totally select 1738 nodules and 1000 non-nodules from 744 CT scans on LIDC-IDRI and 421 nodules from 46 cases on ELCAP. After data augmentation, almost 640 training cases and 160 testing cases were obtained for each nodule type on LIDC-IDRI and achieved 690 testing cases in total on ELCAP. The result showed that the overall classification rate reached 93.1% and 93.9%, respectively, which was promising in clinical practice.

*C. Deep Learning-Aided Pulmonary Nodules Classification Techniques*

Deep learning-aided decision support system, which is kind of knowledge-based end-to-end learning machine can bring about diagnostic conclusions directly from the images without relying on hand-crafted features which are required and indispensable for feature-based methods. In spite of eliminating the need for hand-crafted features, the end-to-end learning machines perform extremely well for pulmonary nodules classification tasks. To date, CNN is the dominant type of end-to-end deep learning machines. Besides, other types of end–to-end learning machines (e.g., multi-tasking learning/transfer learning, deep belief network) have recently been emerging as credible tools for this task.

*1) Two-dimensional (2D) Convolutional Neural Networks*: Tu et al. (2017) implemented classification to automatically categorize solid, part-solid and non-solid pulmonary nodules in CT scans by directly using the hierarchical features learned from the CNN. They conducted a comprehensive performance comparison between the CNN-based categorization method and the histogram-based approach (HIST). Tested on LIDC date set, the CNN model outperformed the HIST method in the classification task. The accuracy of the CNN model (91.7 ± 1.3) was higher than the HIST analysis (73.2 ± 1.8). For precision and sensitivity metrics, the CNN also outperformed the HIST based schemes.

Shen et al. (2017) addressed the lung nodule malignancy classification issue for CT images by proposing a multi-crop CNN (MC-CNN). Instead of using the traditional max-pooling strategy which was a successive operation after convolutional operation, the MC-CNN employed a multi-crop pooling strategy to extract multi-scale features. In particular, the inputs of multi-crop pooling operation were the convolutional features obtained from either the original image or the pooled features. Then the max-pooling was applied on the multi-scale features several times to extract the nodule's information. Two baseline comparison algorithms were implemented, i.e., an autoencoder-based method and support vector machine. Experimental results based on the LIDC-IDRI date set demonstrated that the MC-CNN achieved promising results with the classification accuracy of 87.14% and the AUC score of 0.93.

Sun et al. (2017) compared the performances of three deep learning algorithms, i.e., LeCun model-based CNN (Lecun et al. 1998), deep belief network (Hinton et al. 2006), and stacked denoising autoencoder, with traditional CAD systems using hand-crafted features including density (i.e., average intensity, standard deviation, and entropy), texture (i.e., GLCM, wavelet, LBP and SIFT) and morphological features (area, circularity, and ratio of semi-axis). Based on the LIDC 1018 cases, experiments showed that the CNN achieved AUC of 0.899 ± 0.018, which was higher than traditional hand-crafted features-based algorithms.

Ciompi et al. (2017) proposed multi-stream multi-scale convolutional neural network where nine streams were evenly grouped into three sets and the three streams in one set were fed with a set of triplets of patches (axial, coronal and sagittal view), respectively. Different sets of streams processed the same triplets of patches at different scales, 10 mm, 20 mm, and 40 mm, respectively. A soft-max layer with six neurons computed the probability for six classes, i.e., the solid, non-solid, part-solid, calcified, perifissural and spiculated type. The proposed network was trained by Multicentric Italian Lung Detection (MILD) trial data (1805 nodules from 943 subjects) (Pastorino et al. 2012; Sverzellati et al. 2016) and validated on Danish Lung Cancer Screening (DLCST) Trial data (639 nodules from 468 subjects) (Pedersen et al. 2009). The results showed that the proposed deep learning network performed better in classification of solid, calcified and non-solid nodules than in the classification of part-solid, perifissural and spiculated nodules in terms of the precision and sensitivity. The reason laying behind the perform difference is that the samples for part-solid, perifissural and spiculated nodules are relatively smaller, which may be compensated in the future by adding more training samples for these categories. Compared with two support vector machines with intensity features and *k*-means unsupervised learning features, the multi-stream multi-scale CNN was much better than two SVMs, with accuracy of 79.5% versus 27% and 39.9%.

Liu et al. (2018) proposed multi-view multi-scale CNNs for classifying nodule type in CT images. In the proposed CNNs, first, a view independent CNN model was pre-trained by captured nodules at sorted scales and views, and then a multi-view CNN model was trained by maximum pooling. Not only solid nodule types such as circumscribed and vascularized, but also ground glass optical and non-nodule types can be discriminated. For the almost 1000 cases of LIDC-IDRI and 690 cases of ELCAP (Early Lung Cancer Action

Project), the overall classification rate reached 92.1% and 90.3%, respectively.

*2) Three-dimensional (3D) Convolutional Neural Network*: To take advantage of three-dimensional spatial contextual information of pulmonary nodules, Liu et al. (2017) developed three-dimensional (3D) CNNs where 3D convolutional layers and cubic input images were used. Tested on the date set containing 326 nodules from the National Lung Cancer Screening Trial (NLST) (The National Lung Screening Trial Research Team 2011) and Early Lung Cancer Action Program (ELCAP), both the 3D CNN single model (AUC of 0.732) and the ensemble models with 3D CNNs (AUC of 0.780) achieved satisfied performance.

*3) Transfer Learning*: Much progress achieved by deep learning for diagnosis of pulmonary nodules has been made using supervised learning schemes which are trained to replicate/imitate the decisions of human experts. Nevertheless, it is very expensive, time-consuming and unreliable to set up large labeled date set through experienced radiologist. Moreover, small amount of training samples based supervised learning schemes leads to overfitting and convergence issues (Yang et al. 2018). There are pressing needs to develop and design novel diagnosing models and algorithms which could achieve super-physician competences from an initial state of tabula rasa, through novel mechanisms beyond the currently prevailing supervised learning scheme. Among them are the multi-tasking learning/transfer learning and unsupervised learning.

In this regard, Hussein et al. (2018) addressed the malignancy of lung nodules through supervised and unsupervised learning approaches. For supervised schemes, a 3D CNN and graph regularized sparse multi-task learning (MTL) were proposed. In particular, the 3D CNN was pre-trained on large-scale sports video which had 487 classes. Then the pre-trained 3D CNN was fine-tuned to learn the discriminative features using samples from lung nodule date set. Finally, the learned attributes were fused in graph regularized sparse multi-task learning. It was the first attempt to employ the transfer learning to 3D CNN to enhance the pulmonary nodule characterization. Moreover, to mitigate the limited availability of labeled training data, unsupervised learning scheme was also investigated by combining clustering and proportion-SVM for malignancy determination. In this hybrid unsupervised learning, the clustering was employed to obtain an initial set of labels, and then the label proportions obtained from clustering were to train the proportion-SVM. Tested on the LIDC-IDRI date set, the proposed graph regularized MTL (91.26%) outperforms the 3D CNN MTL with Trace norm (80.08%) and GIST features with LASSO (76.83%) with a significant margin in terms of the accuracy.

Nishio et al. (2018b) compared performances of conventional CAD, CAD architectured by deep CNN with and without transfer learning. A total of 1236 nodules were selected for evaluation, including 412 benign nodules, 571 primary lung cancer, and 253 metastatic lung cancer. The conventional CAD extracted features through local binary pattern and then fed the features to SVM for classification tasks. The deep CNN was modified by VGG-16. For transfer learning, CNN was pretrained using ImageNet. The result showed that the validation accuracy of CNN with transfer learning achieved 68.0%, which was better than CNN without transfer learning and conventional CAD.

*4) Deep Belief Network*: As kind of unsupervised learning scheme, Restricted Boltzmann Machines (RBM) is trained to learn a feature description directly from the training data with respect to a generative learning criterion. RBM can learn representations from unlabeled data and may outperform standard filter banks (Hinton and Salakhutdinov 2006; Smolensky 1986; van Tulder and de Bruijne 2016). Hua et al. (2015) proposed deep belief network (DBN) for malignant and benign classification in CT image. The DBN was established by constructing stacked RBMs iteratively with three hidden layers and a visible layer. Tested on the LIDC date set, DBN (sensitivity of 73.4% with specificity of 82.2%) and deep CNN (sensitivity of 73.3% with specificity of 78.7%) outperformed *k*-nearest neighbors with SIFT and LBP (sensitivity of 75.6 with specificity of 66.8%), and support vector machine with fractal analysis (sensitivity of 50.2% with specificity of 57.2%).

*D. Comparisons of Pulmonary Nodules Classification Algorithms*

It has been witnessed from aforementioned investigation that significant and remarkable advances have been achieved to deal with the grand challenges on how to precisely differentiate between benign and malignant nodules. Considerable efforts have been expended to advance the computer-assisted nodules classification techniques from the ad hoc conventional approaches to the knowledge-based end to end deep learning techniques. The evolving path is depicted in Fig. 6. To further demonstrate the performance differences, we compare the classical methods and deep learning aided methods in terms of accuracy, specificity, sensitivity, and AUC, which is demonstrated in Table 4.

**Table 4** Comparison between traditional machine learning and deep-learning method for nodule classification

| *Methods* | | *Authors* | *Accuracy* | *Specificity* | *Sensitivity* | *AUC* | *Data set* |
|---|---|---|---|---|---|---|---|
| Traditional Machine Learning | SIFT+LBP+KNN | Hua et al. (2015) | -[b] | 66.80% | 75.60% | - | Authors' compiled data set |
| | Fractional analysis + SVM | Hua et al. (2015) | - | 57.20% | 50.20% | - | Authors' compiled data set |
| | 3D LBP + SVM +TPE | Nishio et al.[a] (2018a) | 79.7% | - | - | 0.850 | LUNGx challenge and NSCLC |
| | Clustering + SVM | Hussein et al. (2018) | 78.06% | 78.28% | 77.85% | - | LIDC-IDRI |
| | MTANN | Suzuki et al. (2005) | 56.4% | 48.4% | 100% | 0.8820 | Authors' compiled data set |
| | MTANN | Tajbakhsh and Suzuki (2017) | - | - | - | 0.8806 | Authors' compiled data set |
| | 3D LBP + XGBoost + TPE | Nishio et al.[a] (2018a) | 82.0% | - | - | 0.896 | LUNGx challenge and NSCLC |
| Deep Learning | Fused features (texture, shape, and deep features) + | Xie et al. (2018) | 89.53%, 87.74%, and 71.93% | 92.02%, 89.67%, and 84.85% | 84.19%, 81.11%, and 59.22% | 0.9665, 0.9445, and 0.8124 | LIDC-IDRI (nodules with malignancy rate 3 |
| | AdaBoost BPNN | | | | | | were discarded / regarded as benign / regarded as malignant) |
| | CNN+SVM | Yuan et al. (2018) [c] | 93.1% | - | - | - | LIDC-IDRI, |
| | | | 93.9% | | | | ELCAP |
| | 2D CNN | Tu et al. (2017) [c] | 91.7% ± 1.3% | - | - | - | LIDC |
| | | Shen et al. (2017) | 87.14% | - | - | 0.93 | LIDC-IDRI |
| | | Hua et al. (2015) | - | 78.70% | 73.30% | - | LIDC |
| | | Ciompi et al. (2017) [c] | 79.50% | - | - | - | MILD and DLCST |
| | | Liu et al. (2018) [c] | 92.1% | - | - | - | LIDC-IDRI |
| | | | 90.3% | | | | ELCAP |
| | | Sun et al. (2017) | - | - | - | 0.899±0.018 | LIDC |
| | DBN | Hua et al. (2015) | - | 82.20% | 73.40% | - | LIDC |
| | 3D CNN | Liu et al. (2017) | - | - | - | 0.732 | NLST and ELCAP |
| | 2D CNN + transfer learning | Nishio et al. (2018b) | 68.0% | - | - | - | Authors' compiled data set |
| | 3D CNN + transfer learning | Hussein et al. (2018) | 91.26% | - | - | - | LIDC-IDRI |

[a] Nishio et al. used SVM and XGBoost for nodule classification, which were separately listed in two entries

[b] The sign "-" means missing data

[c] These paper classified nodule types rather than classifying malignant-benign nodules. Accuracy was measured on all types of nodules

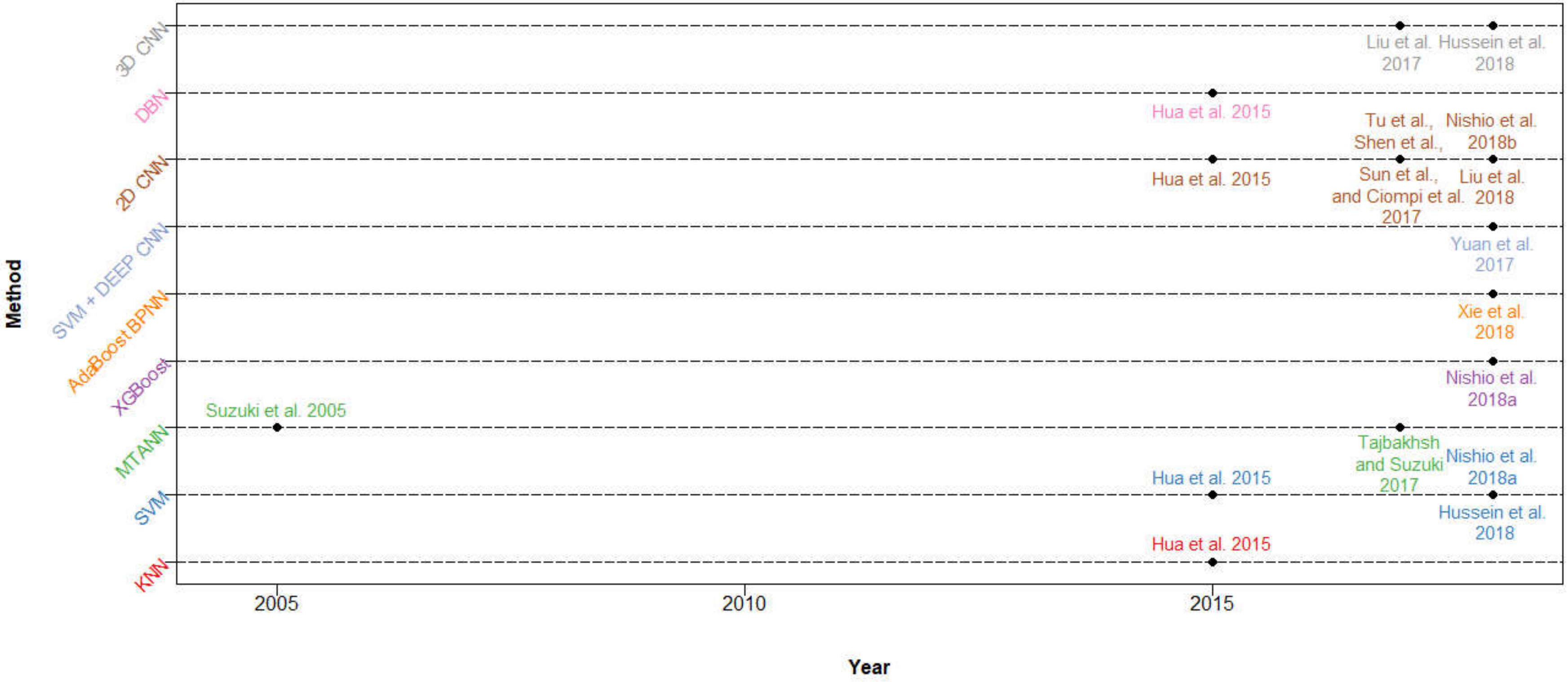

**Fig. 6** Evolving path of computer-assisted benign-malignant pulmonary nodules classification techniques from ad hoc conventional approaches to the knowledge-based end-to-end deep learning techniques

## IV. Challenges and Future Developments

Despite of the emerging of artificial intelligence, especially deep learning as promising decision supporting approaches to automatically analyze chest images for diagnosing purposes, the promising AI and deep learning techniques raise daunting challenges from data scarcity, training efficacy, and diagnostic accuracy. These challenges introduce several opportunities and future research possibilities. In this section, we indentify the challenges and highlight opportunities for future work.

### *A. Data set Scarcity*

One of the crucial issues which hinder the prevailing of the deep learning in medical image analysis is the scarcity of high-quality labeled date set. To effectively train a deep learning model in which tons of parameters need to be estimated, a large volume of labeled data is required. There are thus pressing needs for setting up large-scale date set for the purpose of diagnosing of pulmonary nodules, just as the ImageNet (Russakovsky et al. 2015) which addresses the imposing challenges arising from the natural image analysis. However, differing significantly from the ImageNet for general purpose, it needs everlasting endeavors, tremendous efforts, and sophistication in evaluation of disease patterns by the professionals to build up the date set for medical purpose. Traditionally, manipulation of labeled date set leads to an enlarged volume of labeled data. Lo et al. generated multi-angle patches containing the regions of interest in Cartesian coordinate system (Lo et al. 1993b). Rather than the widely used Cartesian coordinate system, Yang et al. suggested that rotation of the patches containing the regions of interest by small angles could generate more labeled data under spherical and cylindrical coordinate systems (Yang et al. 2018). Ciompi et al. (2017) expanded Multicentric Italian Lung Detection (MILD) trail date set from 1805 nodules (943 subjects) to about 500,000 training samples by rotating the extracted patches, and shifting nodule center randomly within a sphere of 1 mm radius.

*1) Semi/un-supervised Learning*: Semi-supervised learning and unsupervised features learning could be adopted to automatically extract the pulmonary nodule features from the very limited number of annotated medical images, and perform labeling work on the large amount of unlabeled data without the need of intensive physician labeling labor. As observed that in medical image analysis high quality labeled date set is very scarce, e.g., 104 PET/CT images (Teramoto et al. 2016), and only 1018 cases in LIDC (Armato et al. 2010; Armato et al. 2011), effective semi/un-supervised learning techniques could be developed to perform feature extraction on the above small amount of labeled data in a supervised way, then annotate the unlabeled data through unsupervised learning. Hinton (2018) stated that by mimicking human cognition in finding structure in unlabeled data, unsupervised learning could create a smorgasbord of complex feature detectors based on unlabeled data. One work in this direction could be referred to Chen et al. (2018). Another work can be referred to Hussein et al. (2018). In their work, to mitigate the limited availability of labeled training data, unsupervised learning scheme was investigated in which the clustering was employed to obtain an initial set of labels, and then the label proportions obtained from clustering were to train the proportion-SVM.

*2) Transfer Leaning and Multi-task Learning*: The leveraged knowledge from transfer learning among different diseases' patterns may have potential benefits in diagnosis of single disease. The intrinsic learning mechanism underlying deep CNN requires a large amount of labeled data to perform supervised learning. Since the scarcity of publicly available medical images, one of the central questions in deep learning for medical analysis is: Can pre-trained deep CNN with fine-tuning using for instance, a large set of labeled natural images outperform, or in the worst case, perform as well as the CNN trained using medical images from scratch? Tajbakhsh et al. (2016) addressed the issue of knowledge transfer between natural and medical images, between which substantial differences exist. Their experiments suggested that the use of a pre-trained CNN with adequate fine-tuning outperformed or, in the worst case, performed as well as a CNN trained from scratch. It may be taken a consideration of knowledge transfer learning among different diseases in medical images for future study. Nishio et al. (2018b) experimented with pretrained CNN using ImageNet, which was transferred to 2D lung nodule classification task. The result showed that CNN with transfer learning has slightly higher accuracy comparing with CNN without transfer learning. Moreover, employment of the transfer learning to 3D CNN would be highly beneficial and some steps in this direction can be seen in Hussein et al. (2018).

*3) Crowd sourcing to Build Ground-truth Labeled Date set*: It has been observed that large-scaled date sets for general purpose have been annotated though crowd sourcing (i.e., outsourcing of the annotation tasks to a variety of competent sources) to generate ground-truth labeled ones. In healthcare, to date, though crowd sourcing has shown high potentials, such as optimizing treatment plans, predicting disease outbreaks, patient centered price transparency (Meisel et al. 2016) and drug discovery (Lessl et al. 2011), few work is initiated to annotate the medical images by crowd sourcing since outsourcing of the medical annotation work to non-expertise would lead to misclassifications (Chartrand et al. 2017; Weese and Lorenz 2016; Zygmont et al. 2016). Some steps in this direction could be referred to the work by Albarqouni et al. (2016). They developed deep learning networks to generate a ground-truth labeling from non-expert crowd annotation for detection of mitosis in breast cancer histology images, and they showed that outsourcing of the annotation task to the crowd possessing nonprofessional knowledge can perform as well as to the professionals. One interesting question going forward is if the convolutional neural network via additional crowd sourcing layer (AggNet) by Albarqouni et al. (2016) can perform well on other medical tasks, such as pulmonary nodules diagnosis. Granted, privacy problem could be one major obstacle for mass data collection to better train CADs, and in several countries policies were also issued upon personal information. He et al. (2019) discussed the General Data Protection and Regulation (GDPR) issued by European Union (EU) which became effective since May 2018. The GDPR requires informed consent from patients before the collection of any kind of personal information. In addition, it also allows individuals whose data were collected to track the usage of their information and to remove their data at any time. These requirements affect crowd sourcing and massive AI implementation. However, crowd sourcing still remains promising in long term. As He et al. (2019) suggests, GDPR may slow down mass AI implementation in short term as it sets higher bars for data collection, while in long term this kind of regulations may increase credibility of AI implementation, and promote patient participation into crowd sourcing.

*B. Training Efficacy*

Regardless of the negative impact from scarcity of labeled training data on training and generalization performances, either training deep network from scratch or pre-training deep network plus fine training requires iterative adjustment of the network parameters through optimization, and selection of subset of date set for fitting through sampling, leading to time-consuming training process as well as suffering from slow, premature or non convergence, even with the aid of GPU speedup. To address the time-consuming and slow-convergence issues, we highlight several potential directions, including optimal sampling strategy, sophisticated optimization techniques on complex high-dimensional landscape, and multi-task Bayesian optimization for optimizing the hyper-parameters.

*1) Optimal Sampling Strategy*: Optimal sampling strategy in line with the distribution assumptions of classes could be developed to learn from the imbalanced date set to achieve precise classifier. Sampling for learning from the imbalanced date set is a challenge. In medical image analysis, usually the date sets are composed of normal and abnormal cases within which subcategories exist. For instance, pathology patterns of interstitial lung diseases include micronodules, ground glass opacity, reticulation, honeycombing, and consolidation (Anthimopoulos et al. 2016), and PanCan model of nodule type includes solid, non-solid, part-solid, calcified, perifissural and spiculated type (Ciompi et al. 2017). The classification categories between normal and abnormal cases, as well as the subcategories within abnormal examples are not approximately uniformly represented, leading to imbalanced date set. Treating the imbalanced date set equally results in unbalanced allocation of computational budgets between the majority and the minority classes, that is, over-sampling of the majority class while under-sampling of the minority class, consequently leading to degeneration of the efficiency of the learning process and increase of the training time. The concern for learning from the imbalanced date set focuses on deliberated balance between over-sampling of the minority class and under-sampling of the majority class to achieve better classification performance. Several work in this direction could be referred to van Grinsven et al. (2016), Batista et al. (2004) and Chawla et al. (2002).

*2) Sophisticated Optimization on Complex High-dimensional Landscape*: Training deep network is essentially to estimate its parameters/weights. From the viewpoint of optimization, it could be formulated as a very large-scale multi-modal constrained optimization problem with huge amount of decision variables (i.e., network parameters). The purpose of optimization is to search the (near)-optimal or feasible parameter settings for the deep network which fit the sample date set well with respect to certain criteria, e.g., maximization of classifier performance in terms of AUC, Receiver Operating Characteristic (ROC) convex hull, etc. The objective function surface could be seen as landscape, and manipulating the network parameters (i.e., find optimal parameter settings) could be seen as the guided walk on the landscape to climb to the highest point in the landscape for maximization optimization or to lowest point for minimization optimization (Van Cleve and Weissman 2015; Weinberger 1990; Wright 1932). Its landscape is usually characterized by high-dimensional, rugged, strong basin of attraction, huge number of local optima, unknown distribution of the local optima, and large flat plateaus, which make the search extremely difficult. The extensively used stochastic gradient descent optimization method can quickly find a good set of weights, however, it is sensitive to the initial guess and risks getting prematurely stuck in local optima (LeCun et al. 2015; Liu et al. 2011). Given these difficulties, sophisticated optimization methods of high efficacy should be proposed which are capable of on the one hand exploring the high-dimensional complex vast landscape efficiently to find the promising regions, and on the other hand finely exploiting the selected regions with potential impetus to escape from the basins of attraction of the local optima.

Apart from seeking for optimal settings of parameters/weights in the deep network, the proper setting of high-level hyper-parameters in deep learning, e.g., the regularization weights and learning rates, etc., is vital for improving generalization as well as speeding up convergence. Recently, Bayesian optimization yielded state-of-the-art performance in tuning the hyper-parameters of learning models.

Besides, the multi-task Bayesian optimization which could optimize multiple tasks simultaneously by borrowing the multi-task Gaussian process models to Bayesian optimization, could significantly speed up the optimization process when compared to the single-task Bayesian optimization approach (Swersky et al. 2013). One promising direction could be investigating the Bayesian optimization and multi-task Bayesian optimization for optimizing the hyper-parameters in deep network for medical imaging. Some steps in this direction could be referred to the work by Nishio et al. (2018a).

*C. Diagnostic Accuracy*

*1) Explainable Decision to Support Robust and Faithful Diagnosis*: The inference procedure of deep learning is always considered as a black-box process which to date could provide quantitative and accurate decision or predictions. In the sensitive healthcare applications, medical decisions are extremely serious and inappropriate diagnosis may incur improper treatment and legal issues (Charles 2018; Ching et al. 2018). It is imperative that the current black-box-based deep learning system is required to possess the capacity of causal inference, to be specific, the explainable decision, leading to creditable and faithful diagnosis and being invulnerable to adversarial attacks (Finlayson et al. 2018). To fulfill this purpose, the deep learning systems are to identify the full sets of inputs that are responsible for a particular output, and provide answers for "what-if" questions, e.g., what would happen to the diagnosis if some minor perturbations are performed on those features, or what other mechanisms or inputs could lead to the same output (Stoica et al. 2017). Taking pulmonary nodules diagnosis as an instance, if a pulmonary nodule is classified as malignant by analyzing images via deep learning, then it is beneficial to determine which features (e.g., intensity, texture, morphology, spatial contexture, and metabolic features) are explanatory/dependent variable(s) leading to this diagnosis, and what are the impacts of perturbations of those features on the change of diagnosis. Moreover, if the above diagnosis was misclassified, closely related historical data which lead to similar outcome should be investigated to identify the cause of the false diagnosis. Kawagishi et al (2018) proposed an automatic searching method for feature-diagnosis network using Markov chain Monte Carlo (MCMC), after which the conditional probability for each feature could be inferred.

*2) Continual Learning and Shared Learning to Recognize Unprecedented Disease Patterns*: In clinical practice, it usually requires the medical diagnosis system to automatically make decisions in response to unexpected cases. For example, consider unprecedented disease patterns in pulmonary image for diagnosis. Once rare patterns or mixed diseases' patterns are met, the learning system must update their inference procedures and arrive at appropriate diagnostic judgments. However, to date, the learning systems are mostly trained offline and make predictions online, which suggests they can only act in statistic environments not in dynamic ones and cannot handle safely the cases that have not been trained to recognize. In this regard, without retraining the system by incorporating the rare/unprecedented cases, it would continuously make the same wrong inferences. Furthermore, retraining the system by the new case would lead to catastrophic forgetting which means the knowledge the system have already learned would be disrupted (McCloskey and Cohen 1989). These challenges require the learning system to support continual learning to adapt in the changing environment (Thrun 1998). Thus, it could be greatly beneficial to seek to build continual learning systems to support faithful diagnostic decision in the light of dynamic changes by utilizing knowledge already learned to unprecedented disease patterns, while avoiding the effect of catastrophic forgetting (McCloskey and Cohen 1989). One possible learning framework could be Evolved Plastic Artificial Neural Networks (EPANNs) which possess lifelong learning ability to learn and change in response to unseen conditions, and improve with experience (Soltoggio et al. 2018).

Apart from continual learning which is dedicated to make learning system adaptable to unprecedented disease patterns, shared learning on confidential/private data is expected to be beneficial by sharing data among different health care organizations to build complementary date set for training a shared model meanwhile protecting the privacy. However, how to design incentive mechanism to make the competing health care parties to contribute data to enable cooperatively shared learning is a big challenge. For instance, cooperative game theory could be assisted to guide the designing of incentive mechanisms to guarantee that the competing health care parties make more profits (e.g., higher detection rate, less false positive rate) by sharing data than by not sharing data.

*3) Integration with Health Informatics to Enrich Features for Better Diagnosis*: The diagnoses involve very complicated and sophisticated decision making procedures. To improve the accuracy and efficacy of images based diagnoses, comprehensive/related data in the domain of health informatics could be integrated (Ravì et al. 2017). Besides the various formats of medical images, different types of health informatics data have been rising, including electronic health record, gene expression, microRNA, lab tests, wearable device data, social media data (demographic information), cell phone metadata (lifestyle diseases), air pollution prediction data, etc. Development of comprehensive computer-assisted diagnosis schemes would enhance the quality of patient diagnosis by integration of the various types of health informatics data with the medical images. And deep learning is highly adaptable for integrative analysis of heterogeneous data from diverse data sources (Naylor 2018). There are thus pressing needs for proposing novel data mining approaches that can address the imposing challenges arising from large scaled, unstructured, multimodal, heterogeneous, poorly labeled data in health informatics (Müller and Unay 2017). Just as commented in the survey by Miotto et al. (2017), to date there are no researches deploying deep learning technique to combine and mine neither all of these data sources (i.e., electronic health records, imaging, sensor data and text), nor part of them (e.g. only electronic health records and clinical images/genomics) for joint analysis (Miotto et al. 2017). It could be a possible direction to reduce the high false-positive rates as well as to improve the precision of classification by enriching features from various health informatics.

## V. Conclusions

Lung cancer is the most common cancer worldwide and its incidence and mortality are increasing on a yearly basis. In the year of 2018, 2.1 million people are diagnosed and 1.8 million people pass away. Behind the data, the incidence and mortality vary by country and by population. Almost three quarters of deaths occur in low- and middle-income countries. The disparities in incidence and mortality between countries reflect the degrees of socioeconomic development, which leads to the inequalities in access to faithful medical services for cancer prevention, early diagnosis and treatment. The computer-assisted diagnosis is essential to fight the inequality to increase the level of public health and welfares of human beings, since the knowledge-based end to end technology could be rapidly disseminated to the vulnerable population in low economic burden with effectively transferred expertise's knowledge to support faithful diagnostic decision.

Significant and remarkable advances have been achieved since the 1980s. As a promising decision supporting approach, computer-assisted diagnosis automatically analyzes the chest images for pulmonary nodule diagnostic tasks, nodule detection and benign-malignant classification. Along the track of its technological development over the long time span from the 1980s to 2019, we comprehensively examined the computer-assisted nodules detection and benign-malignant classification techniques, which have evolved from the complicated ad hoc analysis pipeline of conventional approaches to the simplified seamlessly integrated deep learning techniques. Various approaches were covered in this investigation, including the pre-deep learning era approaches (e.g., conventional detection techniques, backpropagation neural networks, convolutional neural networks, associative learning neural networks, massive training artificial neural networks, *k*-nearest neighbors, support vector machine, linear discriminant classifier and GentleBoost classifiers), and the deep learning era techniques (e.g., two/three-dimension, multi-view, multi-stream, multi-scale, multi-tasking deep convolutional neural networks, deep belief network, autoencoder network, and the ensemble methods). To date we are in the very fast moving period of computer-assisted pulmonary nodule diagnosis, we are not quite sure that all landmark researches have been involved. We hope that the goal has been met that we claimed to provide a comprehensive research skeleton of success stories of pulmonary nodules diagnosis.

Though the artificial intelligence techniques, especially the deep learning are entering into the clinical medicine, even the U.S. Food and Drug Administration permitted the marketing of the first artificial intelligence-based medical device to detect certain diabetic retinopathy (Food and Drug Administration 2018), tremendous technical barriers need to be crossed to make the medical decisions output from the black box transparent and explainable so that the physicians could be convinced to use them in clinical practice as reliable assistants. We identified the challenges on data scarcity, training efficacy, and diagnostic accuracy for pulmonary nodule diagnosis. We highlighted future research opportunities in learning models, learning algorithms, and enhancement schemes which will shed light on the challenges. Nevertheless the challenges are multifaceted, leading not to relying on merely single solution but on multidisciplinary solutions. For instance, to address the issues from scarcity of high-quality labeled date set, a package of schemes can be jointly designed, including manipulation of labeled date set, automatic labeling based on semi-supervised learning and unsupervised features learning, and knowledge sharing among different diseases' patterns through transfer learning. To improve the accuracy and reduce the false-positive rates, feature enrichment by integration with health informatics for joint analysis could strengthen and improve the image-based diagnosis quality. To address the time-consuming and slow-convergence issues, optimal sampling strategy and sophisticated optimization techniques could be jointly designed to achieve better classification performance.

We wish the challenges indentified and the research opportunities highlighted are significant for bridging current state to future prospect and satisfying future demand. We hope that many promising and potential multidisciplinary researches could be carried out to raise the state-of-the-art clinical applications as well as to increase both welfares of physicians and patients. We firmly hold the vision that fair access to the reliable, faithful, and affordable computer-assisted diagnosis for early cancer diagnosis would fight the inequalities in incidence and mortality between populations, and save more lives.

**Compliance with Ethical Standards**

**Funding**: BL was funded by Key Research Program of Frontier Sciences, Chinese Academy of Sciences (QYZDB-SSW-SYS020), Major Project to Promote Development of Big Data from National Development and Reform Commission (2016-999999-65-01-000696-01). JH was sponsored by Collaboration Research Project of Guangdong Education Department (GJHZ1006 and 2014KGJHZ010). HL was supported by the Medical and Health Science and Technology Project of Guangzhou Municipal Health Commission (20161A011060), the Science and Technology Planning Project of Guangdong Province (2017A020215110), the Natural Science Foundation of Guangdong Province (2018A030313534).

**Conflict of Interests**: The authors declare that they have no conflict of interest.

**Ethical approval**: This article does not contain any studies with human participants or animals performed by any of the authors.